\documentclass[journal]{IEEEtran}
\usepackage{url} 
\usepackage[numbers,sort&compress]{natbib}
\usepackage{cite}
\usepackage{amsthm,amsmath,amssymb}
\usepackage{mathrsfs}
\usepackage{array}
\usepackage{tikz}
\usepackage[skins]{tcolorbox} 
\tcbuselibrary{breakable} 
\usepackage{booktabs}  
\usepackage{multirow}
\usepackage{graphicx}
\usepackage[table,xcdraw]{xcolor}
\usepackage[linesnumbered, ruled]{algorithm2e}
\usepackage{orcidlink}

\usepackage[capitalize]{cleveref}
\crefname{section}{Sec.}{Secs.}
\Crefname{section}{Section}{Sections}
\Crefname{figure}{Figure}{Figures}
\crefname{figure}{Fig.}{Figs.}
\Crefname{table}{Table}{Tables}
\crefname{table}{Tab.}{Tabs.}
\Crefname{algorithm}{Algorithm}{Algorithms}
\crefname{algorithm}{Alg.}{Algs.}
\Crefname{Equation}{Equation}{Equations}
\crefname{Equation}{Eq.}{Eqs.}
\Crefname{Appendix}{Appendix}{Appendices}
\crefname{Appendix}{App.}{Apps.}

\newcommand{\traj}{\mathcal{J}}

\begin{document}
\title{\texttt{M-STAR}: Multi-Scale Spatiotemporal Autoregression for Human Mobility Modeling}

\author{Yuxiao~Luo\orcidlink{0009-0007-6088-0563},
        Songming~Zhang\orcidlink{0000-0001-7695-5880},
        Sijie~Ruan\orcidlink{0000-0002-4520-7174},~\IEEEmembership{Member,~IEEE},
        Siran~Chen \orcidlink{0009-0009-6783-3270},
        Kang~Liu\orcidlink{0000-0001-7466-4123}, 
        Yang~Xu\orcidlink{0000-0003-3898-022X},
        Yu~Zheng\orcidlink{0009-0002-4515-0080},~\IEEEmembership{Fellow,~IEEE},
        and~Ling~Yin\orcidlink{0000-0002-0262-0655}
\thanks{\emph{Corresponding author: Ling~Yin.}}
\thanks{Y. Luo and Y. Xu are with the Department of Land Surveying and Geo-informatics, The Hong Kong Polytechnic University, Hong Kong, China (e-mail: yuxiao.luo@connect.polyu.hk; yang.ls.xu@polyu.edu.hk).}%
\thanks{Y. Luo, K. Liu, L. Yin, and S. Chen are with the Shenzhen Institutes of Advanced Technology (SIAT), Chinese Academy of Sciences, Shenzhen 518055, Guangdong, China (e-mail: kangliu@siat.ac.cn; yinling@siat.ac.cn; chensiran17@mails.ucas.ac.cn).}%
\thanks{S. Zhang and L. Yin are with the Faculty of Computer Science and Control Engineering, Shenzhen University of Advanced Technology, Shenzhen, Guangdong, China (e-mail: zhangsongming@suat-sz.edu.cn).}%
\thanks{S. Ruan is with the School of Computer Science and Technology, Beijing Institute of Technology, Beijing 100081, China (e-mail: sjruan@bit.edu.cn).}%
\thanks{Y. Zheng is with JD Technology \& JD Intelligent Cities Research, Beijing 100176, China, and also with Beijing Key Laboratory of Traffic Data Mining and Embodied Intelligence, Beijing 100044, China (e-mail: msyuzheng@outlook.com).}%
}


\markboth{Journal of \LaTeX\ Class Files}%
{Luo \MakeLowercase{\textit{et al.}}: Bare Demo of IEEEtran.cls for IEEE Journals}

\maketitle

\begin{abstract}
Modeling human mobility is vital for extensive applications such as transportation planning and epidemic modeling. With the rise of the Artificial Intelligence Generated Content (AIGC) paradigm, recent works explore synthetic trajectory generation using autoregressive and diffusion models. 
While these methods show promise for generating single-day trajectories, they remain limited by inefficiencies in long-term generation (e.g., weekly trajectories) and a lack of explicit spatiotemporal multi-scale modeling.
This study proposes \textbf{M}ulti-Scale \textbf{S}patio-\textbf{T}emporal \textbf{A}uto\textbf{R}egression (\textbf{M-STAR}), a new framework that generates long-term trajectories through a coarse-to-fine spatiotemporal prediction process. M-STAR combines a Multi-scale Spatiotemporal Tokenizer that encodes hierarchical mobility patterns with a Transformer-based decoder for next-scale autoregressive prediction. Experiments on two real-world datasets show that M-STAR outperforms existing methods in fidelity and significantly improves generation speed. The data and codes are available at \url{https://github.com/YuxiaoLuo0013/M-STAR}
\end{abstract}


\section{Introduction}

\IEEEPARstart{S}{patiotemporal} data mining of human mobility has become a foundational component of contemporary urban analytics and management \citep{zheng2015trajectory,shi2020prediction,lin2024unite,lin2023pre}. Mobility data reveal how individuals interact with urban infrastructure and services, providing essential knowledge for modeling collective dynamics and supporting a variety of urban computing tasks. With the advancement of generative artificial intelligence, trajectory generation models have emerged as a novel approach to human mobility modeling \citep{yuan2022activity,zhang2025noise}. These models serve two primary purposes. First, as data surrogates, generative trajectories can replace sensitive individual mobility records without exposing personally identifiable information, thereby alleviating privacy risks and regulatory constraints \citep{jin2022survey,wang2024spatiotemporal}. This substitution supports downstream mobility-driven applications such as traffic prediction \citep{pan2020spatio}, epidemic control \citep{luo2025architecting}, crime risk prediction \citep{wang2024diffcrime}, and urban planning \citep{yuan2014discovering,ruan2022discovering}. Second, as model-based simulation engines, generative models provide a flexible and secure reproduction of intra-city mobility dynamics, functioning as human dynamic engines for urban digital twins \citep{luo2025traveller}.
\begin{figure}[t]
    \centering
    \includegraphics[width=0.99\linewidth]{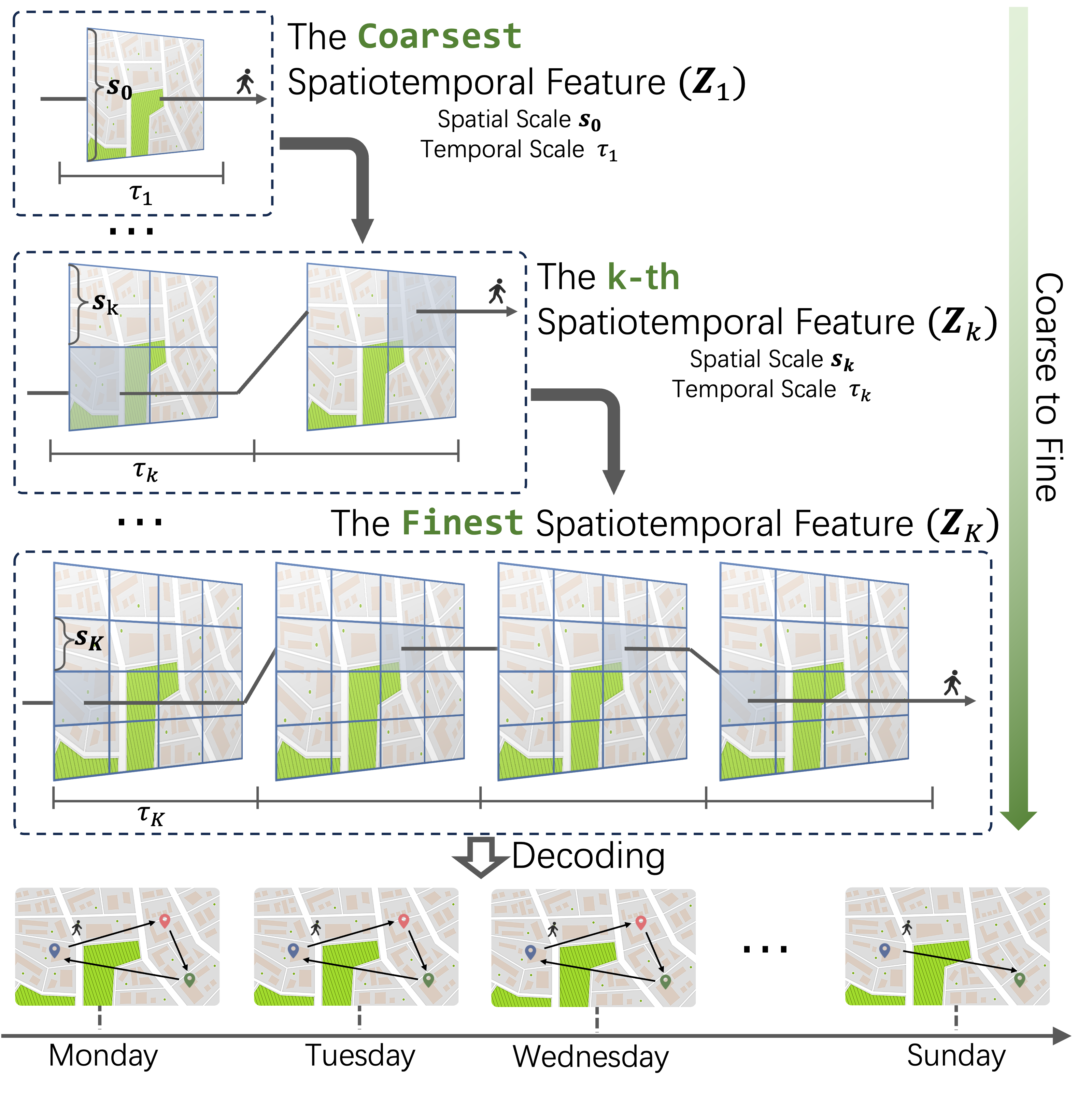}
    \caption{High-level overview of M-STAR’s coarse-to-fine trajectory generation across spatial and temporal scales.}
    \label{fig:intro}
\end{figure}

Autoregressive and diffusion models have recently been introduced to trajectory generation, drawing upon their respective strengths in sequential modeling \citep{vaswani2017attention,zhong2023e2s2} and visual data generation \citep{ma2024follow,tulyakov2018mocogan}. 
Despite their promise on short-term (e.g., one-day) generation, these methods struggle with long-term mobility modeling (e.g., one week), which better aligns with the natural rhythm of human activities.
\emph{\textbf{First}}, autoregressive models \citep{feng2020learning,yuan2022activity,Wang_2024} are prone to error accumulation over extended sequences and often struggle to capture the global structure and fine-grained spatial correlations inherent in human mobility patterns.
\emph{\textbf{Second}}, diffusion models like CoDiffMob \citep{chu2024simulating} are effective at modeling spatial dependencies; however, they require multiple iterative denoising steps, resulting in slow generation speeds that become impractical for long-term trajectory synthesis.
\emph{\textbf{Third}}, although several prediction-oriented models have explored multi-scale representations \citep{zhang2023spatialtemporalinterplayhumanmobility}, most existing trajectory generation methods still lack explicit mechanisms for modeling spatiotemporal patterns across multiple scales. Multi-scale modeling is essential because human mobility exhibits hierarchical and scale-dependent structures in both space and time \citep{tan2025spatiotemporal}.
At the micro scale (short time $\times$ small space), individuals engage in fine-grained movements such as visiting nearby restaurants or making quick neighborhood errands, which require modeling localized spatial variability and minute-level temporal changes.
At the meso scale (hourly $\times$ cross-region), recurring patterns such as daily commuting or school–home travel dominate, reflecting structured inter-regional flows and regular temporal rhythms.
At the macro scale (day–week $\times$ city-wide), mobility is shaped by broader temporal cycles and large-scale spatial organization, such as weekly activity schedules, weekday–weekend differences, and occasional long-distance trips like weekend hiking or inter-district leisure travel.
Capturing these interacting scales is essential for long-term trajectory generation; however, explicit modeling of such multi-scale spatiotemporal effects remains unexplored.

To address these challenges, we propose \textbf{M}ulti-Scale \textbf{S}patio\textbf{T}emporal \textbf{A}uto\textbf{R}egression (\textbf{M-STAR}), a novel framework that generates long-term human trajectories through a coarse-to-fine spatiotemporal prediction process (see \cref{fig:intro}). Unlike conventional token-wise autoregressive models that generate trajectories one location at a time, M-STAR performs autoregression over scale levels, where each autoregressive unit outputs a complete spatiotemporal mapping. This design enables explicit modeling of multi-scale mobility patterns across long-term human mobility and substantially accelerates generation, as the number of scale levels is far smaller than the trajectory length. Specifically, we first use the Multi-Scale Spatiotemporal Tokenizer (\textbf{MST-Tokenizer}) to transform input trajectories into hierarchical representations by projecting them onto spatial grids of varying resolutions and aligning them with temporally downsampled sequences. These multi-scale spatiotemporal embeddings are then compressed into discrete residual token sequences using a residual vector quantization scheme. These residual tokens carry explicit physical meaning: each token represents the information difference between adjacent coarse- and fine-scale representations, capturing how localized mobility patterns deviate from broader regional structures. By encoding these cross-scale residuals, the quantization process preserves both global mobility trends and fine-grained behavioral variations. Finally, conditioned on these tokens, the Spatio-Temporal Autoregressive Trajectory Transformer (\textbf{STAR-Transformer}) performs autoregressive next-scale prediction, progressively generating finer-resolution trajectories by attending to coarser-level contexts and individual mobility attributes. To ensure temporal coherence and sampling diversity, we further introduce a token-wise adaptive temperature sampling strategy and a moving-average smoothing, which together enhance the realism and consistency of the generated trajectories. To the best of our knowledge, M-STAR is the first framework to generate human mobility trajectories using next-scale prediction across progressively refined spatiotemporal scales. In summary, our key contributions can be summarized as follows:
\begin{itemize}
\item We propose M-STAR, a novel framework for generating long-term human mobility trajectories by progressively refining predictions across spatiotemporal scales.
\item We design the MST-Tokenizer to hierarchically encode trajectories via residual vector quantization, followed by the STAR-Transformer, which autoregressively predicts finer-scale tokens conditioned on coarser context and movement attributes.
\item Extensive experiments on real-world datasets demonstrate that M-STAR not only outperforms state-of-the-art baselines in trajectory fidelity but also achieves significantly faster generation speed compared to diffusion-based models.
\end{itemize}

\section{Preliminary}
\begin{figure*}[!tb]
    \centering
    \includegraphics[width=\linewidth]{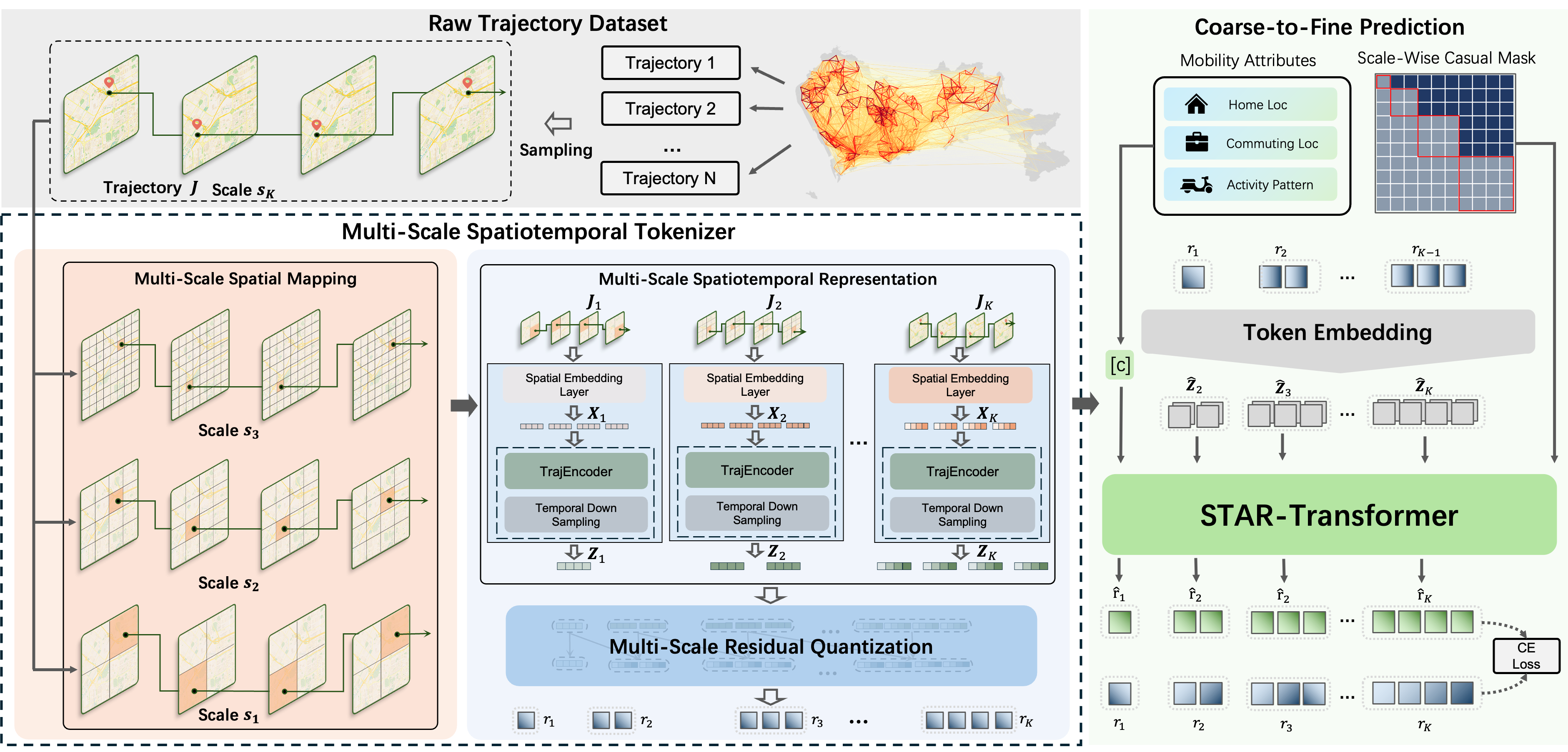}
    \caption{Overview of the M-STAR framework: a coarse-to-fine trajectory generation process that first tokenizes input trajectories into Multi-Scale Spatiotemporal Tokenizer (MST-Tokenizer), followed by autoregressive next-scale prediction using STAR-Transformer.}
    \label{fig:overview}
\end{figure*}
\subsection{Problem Formulation}
We consider a human trajectory generation task in an urban environment, where there are $N$ real-world individual trajectories  $\{ \traj^n\}_{n=1}^N$. 
Each trajectory $\traj^n$ is a sequence of $T$ spatial-temporal points: $[ l_1, l_2, \dots, l_T]$, where $l_t$ denotes the location identifier e.g., grid ID, community ID, and POI ID) recorded from the corresponding temporal slot (e.g., a 1-hour time slot).  We formulate this as a sequence modeling problem, training a generative model $\mathcal{G}$ to align the distribution of generated trajectories with that of real ones
(e.g., a 1-hour time slot).
Specifically, the generation objective is defined as:
\begin{equation}
\mathcal{L} = \ell \left( \{ \mathcal{G}(\traj^n) \}_{n=1}^N, \{ \traj^n \}_{i=n}^N \right),
\end{equation}
where $\ell(\cdot, \cdot)$ measures the distributional similarity of mobility-related statistics (e.g., radius of gyration, waiting time) between real and synthetic trajectories.

\subsection{Formulation of Vector-Quantized Variational Autoencoder}
Vector-Quantized Variational Autoencoder (VQ-VAE) \citep{van2017neural} provides a discrete latent representation by mapping continuous encoder features to the nearest entries in a learnable codebook. Unlike standard VAEs with continuous Gaussian latents, VQ-VAE replaces latent variables with discrete codeword indices, enabling stable training and effective autoregressive modeling.
Given an input sequence, the encoder produces continuous latent vectors $\mathbf{z} \in \mathbb{R}^D$. Each vector is quantized by selecting the nearest codeword from a codebook $\mathcal{Q}$:
\begin{equation}
    q=\underset{v\in[V]}{\operatorname*{\operatorname*{\arg\min}}}\operatorname{dist}\left(\mathbf{z},\mathbf{q}_v\right).
\end{equation}
The quantized latent $\mathbf{q}$ is then passed to the decoder to reconstruct the original data. 

To enable generation, VQ-VAE learns an autoregressive prior over the sequence of discrete latent tokens $\mathbf{q}_{1:K} = (\mathbf{q}_1, \mathbf{q}_2, \ldots, \mathbf{q}_K)$. The prior factorizes the joint distribution as:
\begin{equation}
    p(\mathbf{q}_1, \mathbf{q}_2, \ldots, \mathbf{q}_K)
= \prod_{k=1}^{K} p(\mathbf{q}_k \mid \mathbf{q}_{<k}),
\end{equation}
where each token $\mathbf{q}_k$ is predicted conditioned on all previously generated tokens, and decoding the full token sequence through the VQ-VAE decoder to obtain synthesized data. This discrete autoregressive mechanism provides the foundation for the multi-stage residual quantization and coarse-to-fine generative process employed in the M-STAR framework.

\begin{table}[t]
    \centering
    \caption{Notation used in the M-STAR framework.}
    \label{tab:notation}
    \begin{tabular}{ll}
        \toprule
        \textbf{Symbol} & \textbf{Description} \\
        \midrule
        $\mathcal{J}$ & Individual trajectory of length $T$. \\
        $N$ & Number of trajectories of the total dataset. \\
        $T$ & Trajectory length. \\
        $K$ & Number of spatiotemporal scales. \\
        $s_k$ & Spatial scale at level $k$. \\
        $\tau_k$ & Temporal scale at level $k$. \\
        $T_k$ & Length after temporal downsampling ($T_k = T/\tau_k$). \\
        $E_k$ & Spatial embedding matrix at scale $s_k$. \\
        $\mathbf{L}_k$ & One-hot location matrix at scale $s_k$. \\
        $\mathbf{X}_k$ & Spatial embeddings at scale $s_k$. \\
        $\mathbf{Z}_k^{\text{enc}}$ & Encoded spatiotemporal features. \\
        $\mathbf{Z}_k^{\text{dec}}$ & Decoded spatiotemporal features. \\
        $\hat{\mathbf{Z}}_k$ & Predicted finer-scale features. \\
        $\mathbf{F}_k$ & Residual features at scale $k$. \\
        $\mathcal{Q}$ & Shared VQ codebook of size $V$. \\
        $\mathbf{r}_k$ & Residual token sequence at scale $k$. \\
        $\mathbf{R}$ & All multi-scale token sequences. \\
        $\mathbf{c}_n$ & Movement attributes of user $n$. \\
        $\alpha$ & A hyperparameter in the token-wise sampling. \\
        \bottomrule
    \end{tabular}
\end{table}
\section{Methodology}
To address challenges in long-term dependency modeling and generation efficiency, we propose M-STAR, a coarse-to-fine framework for trajectory generation. Unlike traditional next-token methods, M-STAR predicts at the scale level, with each autoregressive unit representing a spatiotemporal map. As shown in \cref{fig:overview}, M-STAR consists of two key components: the MST-Tokenizer, which encodes trajectories into hierarchical spatial-temporal tokens, and the STAR-Transformer, which autoregressively generates finer-scale trajectories based on coarser-scale contexts and movement attributes. 
 
\subsection{Multi-Scale Spatiotemporal Tokenizer}
Instead of modeling human mobility at a fixed spatiotemporal scale, we introduce a novel Multi-scale Spatiotemporal Tokenizer (MST-Tokenizer) to extract and encode multi-resolution representations. As illustrated in \cref{fig:overview}, our MST-Tokenizer consists of three stages: Multi-Scale Spatial Mapping, Multi-Scale Spatiotemporal Representation, and Multi-Scale Residual Quantization.
\subsubsection{Multi-Scale Spatial Mapping}
Unlike images or temporal signals \citep{tian2024visual, gong2024carp}, trajectories are inherently defined as sequences of discrete spatiotemporal units (e.g., grid cells with timestamps). As a result, popular operations such as convolution or pooling cannot be directly applied to derive multiscale representations.
To address this, we explicitly redefine spatial granularity by mapping each trajectory $\mathcal{J}\in \mathbb{R}^{1 \times T}$ onto a set of grids with different resolutions $\{s_1, \dots, s_{K}\}$, where finest resolution $s_K$ corresponds to the raw $\mathcal{J}$. Each resolution $s_k$ is a reflection of cell size, and each trajectory point is assigned to a cell in the corresponding grid, yielding a discrete token sequence at each scale. These multi-resolution spatial units enable the model to jointly consider coarse-level movement trends (e.g., across districts) and fine-grained behaviors (e.g., between POIs).

\subsubsection{Multi-Scale Spatiotemporal Representation}
To construct multi-scale representations of human trajectories, we jointly model spatial and temporal scales. Given the trajectory $J_k$, we use one-hot encoding to transform it into a matrix $J_k \in \mathbb{R}^{T \times N_k}$. The spatial representation at scale $s_k$ is then computed as:
\begin{equation}
    \mathbf{X}_k = \mathbf{L}_k \cdot E_k, 
\end{equation} 
where each row of $\mathbf{X}_k \in \mathbb{R}^{T \times d}$ represents the spatial embedding at time $t$ for scale $s_k$.  

Then, capturing complex mobility patterns over timescales (e.g., hourly routines or long-duration movements) requires establishing a connection between spatial and temporal scales. For each spatial scale $s_k \in \{s_1, \dots, s_{K} \}$, we therefore assign a corresponding temporal scale $\tau_k \in \{\tau_1,  \dots, \tau_{K} \}$. Specifically, we construct coarse-grained temporal representations by encoding fine-scale inputs and performing scale-aware downsampling. Given the spatial embedding $ \mathbf{X}_k$ at scale $s_k$, we apply the $\operatorname{TrajEncoder}$ of shared-parameters Transformer encoder to encode temporal dependencies, followed by an interpolation operation for temporal downsampling according to scale $\tau_k$:
\begin{equation}
\mathbf{Z}_k =  \operatorname{Interpolate}(\operatorname{TrajEncoder}(\mathbf{X}_k), \tau_k),
\end{equation}
where $\mathbf{Z}_k \in \mathbb{R}^{T_k \times d}$ is the resulting spatiotemporal representation and $T_k = T / \tau_k$. 
The $\operatorname{Interpolate}$ adopts an area-based function~\citep{lam1983spatial} for temporal downsampling, which averages over local regions to produce coarse-grained sequences. This approach ensures smoother temporal abstraction, especially when the target length is not an integer divisor of the input length.

\subsection{Multi-Scale Residual Quantization}
\begin{figure}[!tb]
    \centering
    \includegraphics[width=0.9\linewidth]{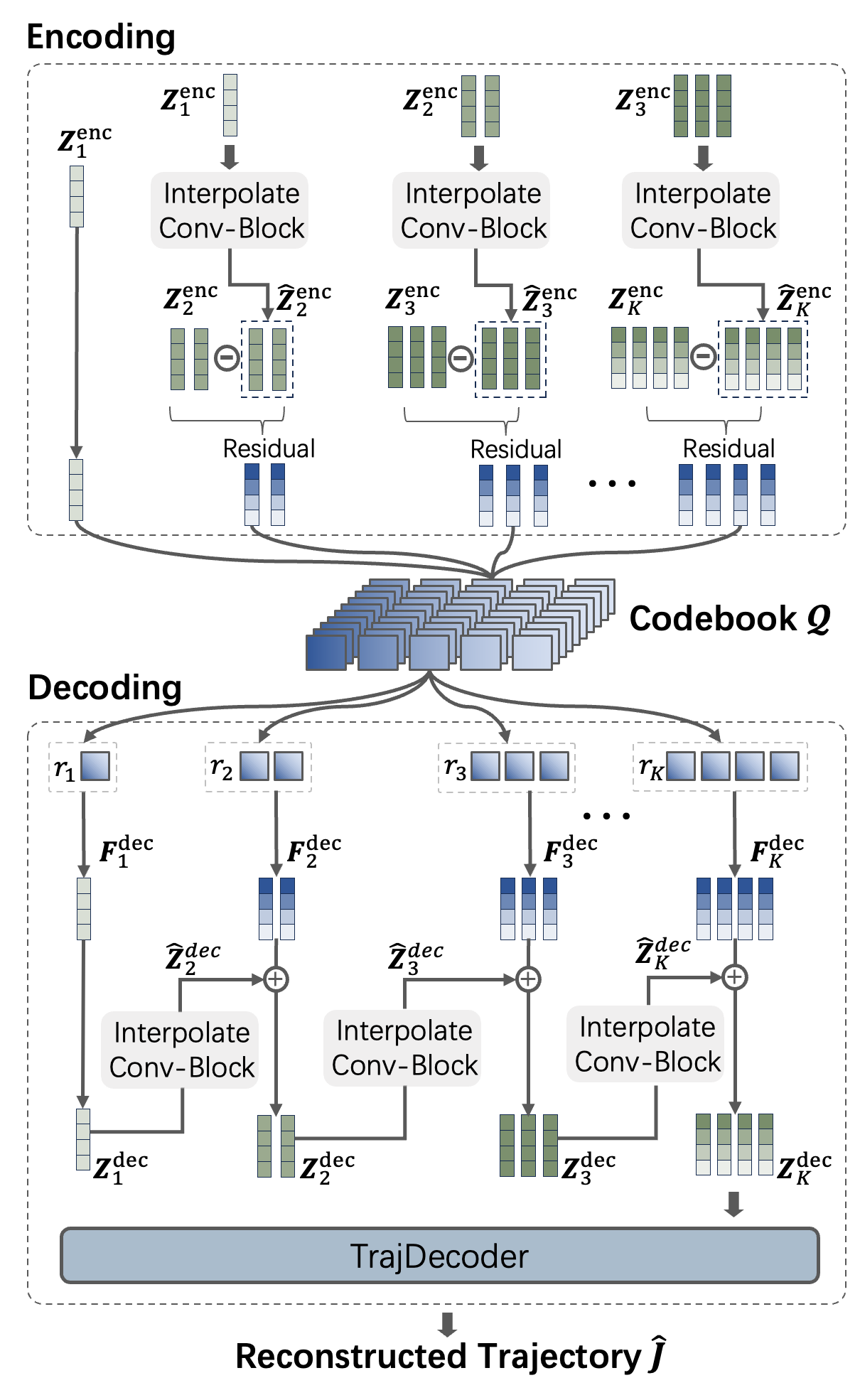}
    \caption{The pipeline of the Multi-Scale Residual Quantization module, illustrating the encoding and decoding phases.}
    \label{fig:quant}
\end{figure}

To convert continuous multi-scale temporal representations $\mathbf{Z}_k$ into discrete token sequences, as shown in \cref{fig:quant}, we employ a residual vector quantization strategy \citep{lee2022autoregressive} based on the VQ-VAE framework \citep{van2017neural}. In this design, the residuals are computed directly between coarse- and fine-scale representations, allowing the model to capture explicit information differences between adjacent scales.
\subsubsection{Encoding Phase}
 We denote the representation at scale $k$ as $\mathbf{Z}_k^{enc}$ for clarity in the encoding phase.
This approach compresses the representation at each scale into a sequence of discrete codeword indices, while progressively refining the approximation through multiple quantization stages.

Given the multi-scale representaion $[\mathbf{Z}_1^{enc}, \dots, \mathbf{Z}_K^{enc}]$, we use the coarser-scale representation $\mathbf{Z}_{k-1}$ to predict the finer-scale feature $\hat{\mathbf{Z}}_k^{enc}$. We then compute the residual between $\hat{\mathbf{Z}}_k^{enc}$ and $\mathbf{Z}_k^{enc}$, which can be formalized as:
\begin{equation}
\begin{split}
\hat{\mathbf{Z}}_k^{enc}&=\operatorname{Conv-Block}(\operatorname{Interpolate}(\mathbf{Z}_{k-1}^{enc},\tau_k)),\\
\mathbf{F}_k^{enc}&=\mathbf{Z}_k^{enc}-\hat{\mathbf{Z}}_k^{enc},
\end{split}
\end{equation}
where $\operatorname{Interpolate}(\mathbf{Z}_{k-1}, \tau_k)$ upsamples the coarser-scale representation $\mathbf{Z}_{k-1}$ to align with the temporal resolution $\tau_k$ of scale $k$. 
The upsampled sequence is then passed through an individual  $\operatorname{Conv-Block}$ for this scale, which consists of a 1D convolution layer followed by a multi-layer perceptron (MLP). This block serves to refine the interpolated sequence and produce a prediction $\widetilde{\mathbf{Z}}_k$ at the finer spatiotemporal scale. 
The resulting residual $\mathbf{F}_k$ captures explicit spatiotemporal meaning, like the difference between regional commuting vs. neighborhood shopping in space and weekly rhythms vs. hourly variations, and reduces redundancy during progressive modeling.

To discretize each residual, we then introduce a shared vector quantizer with a learnable codebook $\mathcal{Q} \in \mathbb{R}^{V \times D}$ across all spatiotemporal scales as the information differences across scales are the same representation type, where $V$ is the number of code vectors and $D$ is the codeword dimension. 
For each feature vector $\mathbf{f} \in \mathbb{R}^{1 \times D}$ in the residual map  $\mathbf{F}_k^{enc}$, the quantizer assigns the nearest codeword index from the shared codebook  $\mathcal{Q}$. This yields a discrete token sequence $\mathbf{r}_k$ at scale $k$:
\begin{equation}
q=\underset{v\in[V]}{\operatorname*{\operatorname*{\arg\min}}}\operatorname{dist}\left(\operatorname{Lookup}(\mathcal{Q},v),\mathbf{f}\right),
\end{equation}
where each token $q$ corresponds to the index of the closest codeword, determined by a distance function $\operatorname{dist}(\cdot)$. Collecting the token sequences from all scales produces the final multi-scale token set $\mathbf{R} = [\mathbf{r}_1, \mathbf{r}_2, \ldots, \mathbf{r}_K]$.

\subsubsection{Decoding Phase}
The decoding process of the final trajectory $\mathcal{\hat{J}}$ follows a similar progressive structure. 
Discrete token sequences $[\mathbf{r}_1, \dots, \mathbf{r}_K]$ are first mapped back to residual features $[\mathbf{F}_1^{dec}, \dots, \mathbf{F}_k^{dec}]$ at corresponding scale $k$ via codebook lookup, and then refined progressively across scales. Finally, the reconstructed feature $\mathbf{Z}_K^{dec}$  at the finest scale is passed to a $\operatorname{TrajDecoder}$ of Transformer architecture for trajectory generation.
\begin{equation}
\begin{split}
\mathbf{F}_k^{dec} &= \operatorname{Lookup}(\mathcal{Q}, \mathbf{r}_k), \\
\hat{\mathbf{Z}}_k^{dec} &= \operatorname{Conv-Block}(\operatorname{Interpolate}(\mathbf{Z}_{k-1}^{dec}, \tau_k)) \\
\mathbf{Z}_k^{dec} &= \hat{\mathbf{Z}}_k^{dec} + \mathbf{F}_k^{dec}, \\
\hat{\mathcal{J}} &= \operatorname{TrajDecoder}(\mathbf{Z}_K^{dec}).
\end{split}
\end{equation}
Here, $\hat{\mathcal{J}} \in \mathbb{R}^{1 \times T}$ denotes the predicted trajectory in discrete grid form. This coarse-to-fine decoding enables effective integration of global movement trends and local behavioral details across all spatiotemporal levels.

\subsubsection{Loss Function} To optimize the tokenizer, we propose a mult-scale VQ-VAE loss $\mathcal{L}_1$ composed of three components: the reconstruction loss, the codebook commitment loss, and the quantization loss. 
The reconstruction loss $\mathcal{L}_{\text{rec}}$ encourages faithful prediction of each location $\mathcal{J}_t$ given the previously generated locations and the finest-scale feature $\mathbf{Z}_K^{dec}$. The commitment loss $\mathcal{L}_{\text{com}}$ ensures that continuous embeddings $\mathbf{Z}_k^{enc}$ remain close to their quantized and decoded counterparts $\mathbf{Z}_k^{dec}$ across all levels. The quantization loss $\mathcal{L}_{\text{qua}}$, applied only at the finest scale by aligning the decoded vector $\mathbf{Z}_K^{dec}$ with $\mathbf{Z}_K^{enc}$, helps prevent gradient accumulation from earlier stages and stabilizes training. 
Thus, the total multi-scale VQ-VAE loss is defined as:
\begin{small}
\begin{equation} 
\setlength{\abovedisplayskip}{3pt}   
\setlength{\belowdisplayskip}{3pt}   
\begin{split}
    \mathcal{L}_1 &= 
\underbrace{- \sum_{t=1}^{T} \log P(\mathcal{J}_t \mid \hat{\mathcal{J}}_{<t}, \hat{\mathbf{Z}}_K)}_{\mathcal{L}_{\text{rec}}} \\
& + \beta \underbrace{\frac{1}{K} \sum_{k=1}^{K} \left\| \operatorname{sg}(\mathbf{Z}_k^{dec}) - \mathbf{Z}_k^{enc} \right\|^2}_{ \mathcal{L}_{\text{com}}}  + \underbrace{\left\| \mathbf{Z}_K^{dec} - \operatorname{sg}(\mathbf{Z}_K^{enc}) \right\|^2}_{\mathcal{L}_{\text{qua}}},
\end{split}
\end{equation}
\end{small}
where $\operatorname{sg}(\cdot)$ is the stop-gradient operator.

\subsection{Trajectory Generation via Coarse-to-Fine Spatiotemporal Prediction}
After obtaining the residual tokens from the MST-Tokenizer, we generate trajectories via STAR-Transformer in a coarse-to-fine spatiotemporal prediction process.  
\subsubsection{ STAR-Transformer}
Unlike conventional next-token prediction with equal-length sequences, next-scale prediction requires generating $T / \tau_k$ finer-scale tokens conditioned on $T/ \tau_{k-1}$ coarser-scale tokens, which differ in length. This scale-wise mismatch poses challenges for parallel training under the teacher-forcing paradigm \citep{lamb2016professor}. For clarity, we illustrate the process using the transition from $\mathbf{r}_{k-1}$ to $\mathbf{r}_k$ as an example.
To address this issue, we do not directly feed $\mathbf{r}_{k-1}$ into the next-scale generator. 
Instead, we first embed the tokens and add them to the representation $\hat{\mathbf{Z}}_{k-1}$, then upsample the sum to the target resolution $\tau_k$, followed by the Conv-Block to obtain $\hat{\mathbf{Z}}_k$ as input.
The generation of the current scale tokens $\mathbf{r}_k$ is then performed via the STAR-Transformer with a block-wise causal attention mask \citep{tian2024visual}:
{ \small \begin{equation}
    \begin{split}
    \hat{\mathbf{F}}_{k-1} &= \operatorname{Lookup}(\mathcal{Q}, \mathbf{r}_{k-1}), \\
    \hat{\mathbf{Z}}_k &= \operatorname{Conv-Block}\left(\operatorname{Interpolate}(\hat{\mathbf{Z}}_{k-1} + \hat{\mathbf{F}}_{k-1}), \tau_k)\right), \\
    \mathbf{\hat{r}}_1,\dots,\mathbf{\hat{r}}_k &= \operatorname{STAR-Transformer}(\mathbf{c}_n,\hat{\mathbf{Z}}_2,\dots,\hat{\mathbf{Z}}_k),
    \end{split}
\end{equation}}
where $\mathbf{c}_n$ represents a start and class token that encodes the $n$-th user's movement attributes, such as home location, commuting location (e.g., workplace or school), and activity-pattern label. Meanwhile, we apply an Adaptive Layer Normalization (AdaLN) module \citep{perez2018film} in each transformer block, enabling the model to modulate intermediate representations based on $\mathbf{c}_n$.

To train the STAR-Transformer for multi-scale residual token generation, we adopt a cross-entropy loss aggregated over all scales:
\begin{equation}
\mathcal{L}_2=\sum_{k=1}^{K}\sum_{i=1}^{T/\tau_k}\left[-\sum_{v=1}^{V}\boldsymbol{r}_{k}^{i,\upsilon}\log \boldsymbol{\hat{r}}_{k}^{i,\upsilon}\right],
\end{equation}
where $V$ is the codebook size, and $T/\tau_k$ denotes the number of tokens at the $k$-th scale. 

During inference, the model predicts $\hat{\mathbf{r}}_k$ conditioned on the previously generated coarser-scale tokens $[\hat{\mathbf{r}}_1, \ldots, \hat{\mathbf{r}}_{k-1}]$ and the start token $\mathbf{c}_n$, which is sampled from the distribution of training dataset and encodes the individual’s movement attributes. These attributes are derived from users’ home and commuting anchors as well as clustered activity patterns, providing high-level behavioral priors for generating coherent trajectories \citep{liu2024act2loc,luo2025traveller}.
The final trajectory $\hat{\mathcal{J}}$ is obtained via $\operatorname{TrajDecoder}(\hat{\mathbf{Z}}_K + \hat{\mathbf{F}}_K)$.
To improve efficiency during autoregressive generation, the STAR-Transformer utilizes key-value (KV) caching to reuse past hidden states, eliminating redundant computation and removing the need for causal masks.

\begin{algorithm}[!tb]
    \caption{Multi-scale Trajectory Generation via STAR-Transformer}
    \label{alg:m-star}
    \SetKwInOut{KwData}{Inputs}
    \SetKwInOut{KwIn}{Hparams}
    \SetKwInOut{KwResult}{Output}
    \KwData{User movement attributes $\mathbf{c}_n$ (home location, commute, etc.)}
    \KwIn{Number of scales $K$, scales $\{(s_k, \tau_k)\}_{k=1}^{K}$, codebook $\mathcal{Q}$}
    
    \tcc{Generate coarsest-scale tokens}
    $\mathbf{r}_1 \gets \operatorname{STAR-Transformer}(\mathbf{c}_n )$\; 
    $\mathbf{F}_1 \gets \operatorname{Lookup}(\mathcal{Q}, \mathbf{r}_1)$\; 
    $\mathbf{Z}_1  \gets \mathbf{F}_1$\; 
    
    \tcc{Autoregressive next-scale generation}
    \For{$k = 2$ \KwTo $K$}{
        $\tilde{\mathbf{Z}}_k \gets \operatorname{Interpolate}(\mathbf{Z}_{k-1} , \tau_k)$\; 
        $\hat{\mathbf{Z}}_k \gets \operatorname{ Conv-Block}(\tilde{\mathbf{Z}}_k)$\; 
        
        $\mathbf{\hat{r}}_k \gets \operatorname{STAR-Transformer}( \mathbf{c}_n ,\hat{\mathbf{Z}}_2,\dots,\hat{\mathbf{Z}}_k)$\; 
        $\mathbf{F}_k \gets \operatorname{Lookup}(\mathcal{Q}, \mathbf{\hat{r}}_k)$\; 
        $\mathbf{Z}_k \gets \hat{\mathbf{Z}}_k + \mathbf{F}_k$\; 
    }
    
    $\hat{\mathcal{J}} = \operatorname{TrajDecoder}(\mathbf{Z}_{K})$\; 
    
    \KwResult{Generated trajectory $\hat{\mathcal{J}}$}
\end{algorithm}

\subsubsection{Dealing with Weaknesses of Next-scale Models}
The proposed next-scale sampling strategy enables autoregressive trajectory generation but introduces two issues:
(1) Intra-scale temporal discontinuity, where independently sampled tokens (e.g., $\Tilde{r}_k^i$ and $\Tilde{r}_k^{i+1}$) may differ even during a stationary period (e.g., staying at home from 8 am to 12 pm with $r_k^i = r_k^{i+1}$), resulting in implausible transitions between different locations during periods of no actual movement.; 
(2) Inter-scale randomness accumulation, where early sampling errors (e.g., incorrectly predicting a workplace instead of a home region at scale $k{-}1$) propagate to finer scales,  causing more errors and noisy trajectories.
To mitigate these two issues, we propose two practical tricks: \emph{1. Token-Wise Sampling.} We assign each token a sampling temperature based on its confidence, computed by normalizing its top-1 logit value across the sequence:
$\text{Temperature}_i = \alpha - \text{confidence}_i$, where $\alpha \geq 1$ is a hyperparameter.
This strategy lowers the temperature for high-confidence tokens (e.g., stay-points) to enhance stability, while raising it for low-confidence tokens (e.g., transitions between locations) to encourage diversity, thereby mitigating temporal discontinuities. \emph{2. Moving-Average Smoothing.} To suppress randomness
 amplification across scales, we apply a smoothing layer based on moving averages over the final generated representation $(\hat{\mathbf{Z}}_K + \hat{\mathbf{F}}_K)$. This reduces high-frequency fluctuations in the output and enhances the temporal coherence of the generated trajectory.

\begin{table}[!t]
\centering
\caption{Statistics of datasets.}
\label{tab:dataset}
\resizebox{0.45\textwidth}{!}{%
\begin{tabular}{lcc}
\toprule
Datasets & Beijing                             & Shenzhen                            \\ \midrule        
\#Duration & Oct 1, 2019 - Dec 31, 2019 & Nov 1, 2021 - Nov 7, 2021 \\   
\#Users &100,000 &200,000 \\
\#Trajectories & 290,836                             & 200,000  \\
\#Girds & 900 & 2,236 \\
\bottomrule
\end{tabular}}
\end{table}

\begin{table*}[!tb]
\centering
\caption{Performance comparison of different trajectory generation methods on long-term trajectories (168 hours). Lower values indicate better performance.
\textbf{Bold} denotes the best result, and \underline{underline} denotes the second-best result.}
\label{tab:main_results}

\resizebox{0.99\textwidth}{!}{%
\begin{tabular}{l|ccccccc|ccccccc}
\toprule
Dataset                  & \multicolumn{7}{c|}{Beijing}                                                                                                                                                       & \multicolumn{7}{c}{Shenzhen}                                                                                                                                                       \\ \midrule
\multirow{2}{*}{Metrics} & \multicolumn{4}{c|}{Individual-level}                                                                   & \multicolumn{2}{c|}{Population-leve}                              & \multirow{2}{*}{\shortstack{Infer. Time\\(Mins)}} & \multicolumn{4}{c|}{Individual-level}                                                                   & \multicolumn{2}{c|}{Population-level}                              & \multirow{2}{*}{\shortstack{Infer. Time\\(Mins)}} \\ \cmidrule(lr){2-7} \cmidrule(lr){9-14}
                         & Displacement    & Radius          & Duration        & \multicolumn{1}{c|}{DailyLoc}          & Flow            & \multicolumn{1}{c|}{Density}      &                              & Displacement    & Radius          & Duration        & \multicolumn{1}{c|}{DailyLoc}          & Flow            & \multicolumn{1}{c|}{Density}      &                              \\ \midrule
W-EPR                    & 0.1527           & 0.3719           & 0.1654           & \multicolumn{1}{c|}{0.6432}           & 1.0291           & \multicolumn{1}{c|}{1.5842}           & -                            & 0.1506          & 0.3483          & 0.2509          & \multicolumn{1}{c|}{0.6477}          & 0.8652          & \multicolumn{1}{c|}{1.1156}          & -                            \\
DITRAS                   & 0.0150          & 0.1821           & 0.0145           & \multicolumn{1}{c|}{0.2314}           & 1.7014           & \multicolumn{1}{c|}{3.4698}           & -                            & 0.0111          & 0.1128          & \underline{ 0.0138}    & \multicolumn{1}{c|}{0.2643}          & 1.9570          & \multicolumn{1}{c|}{3.4200}          & -                            \\ 
Movesim                  & 0.0172          & 0.3547          & 0.0562          & \multicolumn{1}{c|}{0.4734}          & 1.3760          & \multicolumn{1}{c|}{1.3454}          & 16.6                         & 0.1124          & 0.4813          & 0.1293          & \multicolumn{1}{c|}{0.5924}          & 1.0922          & \multicolumn{1}{c|}{2.0967}          & 13.7                         \\
COLA                 & 0.0152          & 0.2126          & 0.0342          & \multicolumn{1}{c|}{0.3225}          & 0.7839          & \multicolumn{1}{c|}{\underline{0.3307}}      & 14.5                         & 0.0158          & 0.3129          & 0.0602          & \multicolumn{1}{c|}{0.3446}          & \underline{0.7621}          & \multicolumn{1}{c|}{\underline{0.3892}}          & 10.5                         \\ 
MIRAGE                & 0.0075          & 0.1274          & 0.0518          & \multicolumn{1}{c|}{0.2025}          & 1.9400          & \multicolumn{1}{c|}{4.5819}     & 46.2                        & 0.0068          & 0.0829          & 0.0587          & 0.1566          & 2.7010         & \multicolumn{1}{c|}{5.5048}      & 31.5                         \\ 
TrajGDM                  & 0.0245          & 0.3395          & 0.0247          & \multicolumn{1}{c|}{0.1660}          & 2.7667          & \multicolumn{1}{c|}{5.8091}          & \textgreater{}100            & 0.0153          & 0.1889          & 0.0218          & \multicolumn{1}{c|}{\underline{ 0.0316}}    & 1.6996          & \multicolumn{1}{c|}{2.1016}          & \textgreater{}100            \\
DiffTraj                 & 0.0156          & 0.0309          & 0.0363          & \multicolumn{1}{c|}{0.0942}          & 1.0098          & \multicolumn{1}{c|}{0.9212}          & 21.8                         & 0.0435          & 0.0390          & 0.1046          & \multicolumn{1}{c|}{0.1107}          & 1.1107          & \multicolumn{1}{c|}{1.5158}          & 15.1                         \\
CoDiffMob                & \textbf{0.0004} & \underline{ 0.0212}    & \underline{ 0.0091}    & \multicolumn{1}{c|}{\underline{ 0.0483}}    & \underline{ 0.7457}    & \multicolumn{1}{c|}{ 0.3936}    & 43.0                         & \underline{ 0.0089}    & \underline{ 0.0297}    & 0.0530          & \multicolumn{1}{c|}{0.0449}          &  0.7855    & \multicolumn{1}{c|}{0.5847}    & 29.7                         \\
\rowcolor[HTML]{EFEFEF} 
M-STAR                     & \underline{ 0.0005}    & \textbf{0.0129} & \textbf{0.0087} & \multicolumn{1}{c|}{\textbf{0.0039}} & \textbf{0.7192} & \multicolumn{1}{c|}{\textbf{0.2403}} & 1.5                          & \textbf{0.0012} & \textbf{0.0197} & \textbf{0.0089} & \multicolumn{1}{c|}{\textbf{0.0047}} & \textbf{0.7312} & \multicolumn{1}{c|}{\textbf{0.3653}} & 1.0\\                          
\bottomrule
\end{tabular}
}
\end{table*}
\section{Experiments}
\subsection{Experimental Settings}
\paragraph{\textbf{Datasets}}
We conduct our experiments on two real-world weekly trajectory datasets from Beijing and Shenzhen, two major Chinese cities, collected by Tencent \citep{shao2024beyond} and China Unicom \footnote{http://www.smartsteps.com}. 
Beijing features a concentric urban layout with relatively stable population flows, while Shenzhen follows a polycentric, belt-like structure along an east-west axis, leading to more diverse and multi-scale mobility behaviors \citep{tan2025spatiotemporal}. Both study areas are divided into 1 km grids, with each weekly trajectory segmented into 168 hourly steps to support long-term mobility generation evaluation. The data is randomly split into training, validation, and test sets with a 5:2:3 ratio. Detailed statistics are shown in Table \ref{tab:dataset}.
\paragraph{\textbf{Baselines and Metrics}}
We compare M-STAR with several state-of-the-art trajectory generation methods, including: two mechanistic models, DITRAS \citep{pappalardo2018data} and W-EPR \citep{wang2019extended}; three autoregression-based models, MoveSim \citep{feng2020learning}, COLA \citep{Wang_2024}, and MIRAGE \citep{deng2025revisiting}; and three diffusion-based models, TrajGDM \citep{chu2024simulating}, DiffTraj \citep{zhu2023difftraj}, and CoDiffMob \citep{zhang2025noise}.
To evaluate the spatiotemporal fidelity of generated trajectories, we adopt metrics from two perspectives:
\textbf{(1) Individual-level metrics:} We use four standard indicators: \textit{Distance} (travel length between consecutive locations), \textit{Radius} (gyration radius around the trajectory center), \textit{Duration} (time staying at each location), and \textit{DailyLoc} (number of distinct visited locations per day) \citep{luca2021survey}. The Jensen-Shannon Divergence (JSD) \citep{menendez1997jensen} measures distributional differences between real and generated data. \textbf{(2) Population-level metrics:} We compute the Mean Absolute Percentage Error (MAPE) for \textit{Flow} (origin-destination movements) and \textit{Density} (population distribution across spatial grids) at each time step. 

In addition to fidelity, we further assess the diversity of generated trajectories by comparing the number of unique trajectories in the generated and real datasets. The \textit{Diversity Error} is defined as:
\begin{equation}
\text{\textit{Diversity Error}} = \frac{N_{gen} - N_{real}}{N_{real}},
\end{equation}
where $N_{gen}$ and $N_{real}$ denote the number of unique trajectories in the generated and real datasets, respectively. A \textit{Diversity Error} closer to zero indicates better preservation of behavioral diversity and reflects the model’s ability to avoid producing overly repetitive or homogeneous trajectories.

\paragraph{\textbf{Implementation Details}}
\label{app:implement}
To balance performance and efficiency, we adopt eight spatiotemporal scales with spatial resolutions $s \in \{1, 1, 2, 2, 4, 4, 8, 8\}$ km and temporal resolutions $\tau \in \{1, 1.5, 2, 3, 6, 12, 24, 168\}$ hours. The MST-Tokenizer uses a codebook size of $V = 4096$ and embedding dimension of 256. The STAR-Transformer is a 4-layer model with sampling parameter $\alpha = 1.3$, chosen via grid search. Both components are trained for 100 epochs with a batch size of 128, using initial learning rates of $1e{-4}$ for MST-Tokenizer and $1\text{e}{-3}$ for STAR-Transformer, with cosine decay. All experiments are implemented in PyTorch, repeated five times on a single NVIDIA RTX A5000 (24GB) GPU.
\begin{figure*}[t]
    \centering
    \includegraphics[width=0.99\linewidth]{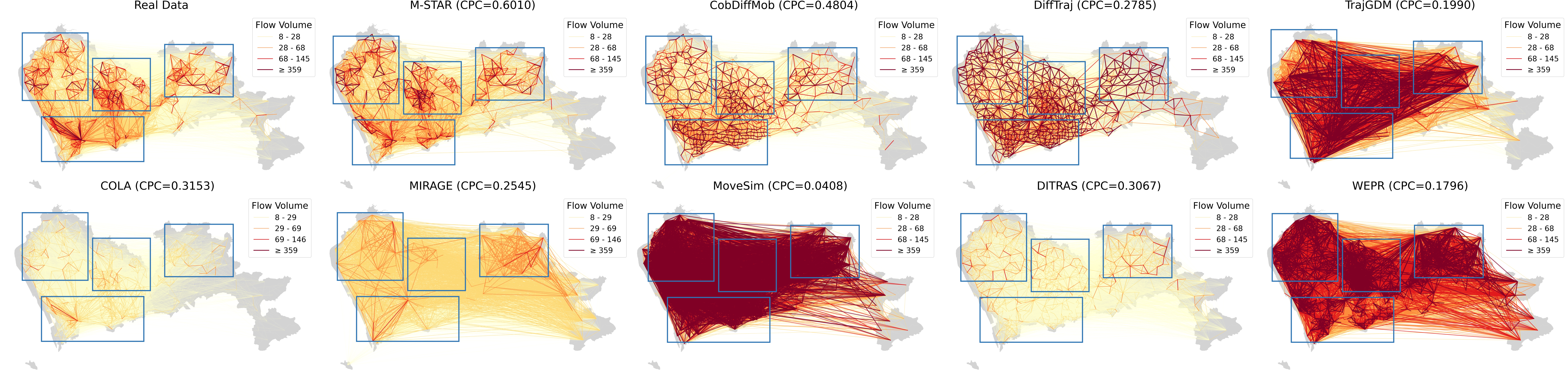}
    \caption{ Visualization of community-level origin-destination flows in Shenzhen}
    \label{fig:od_flow}
    \vspace{-1mm}
\end{figure*}
\begin{figure}[t]
    \centering
    \includegraphics[width=0.99\linewidth]{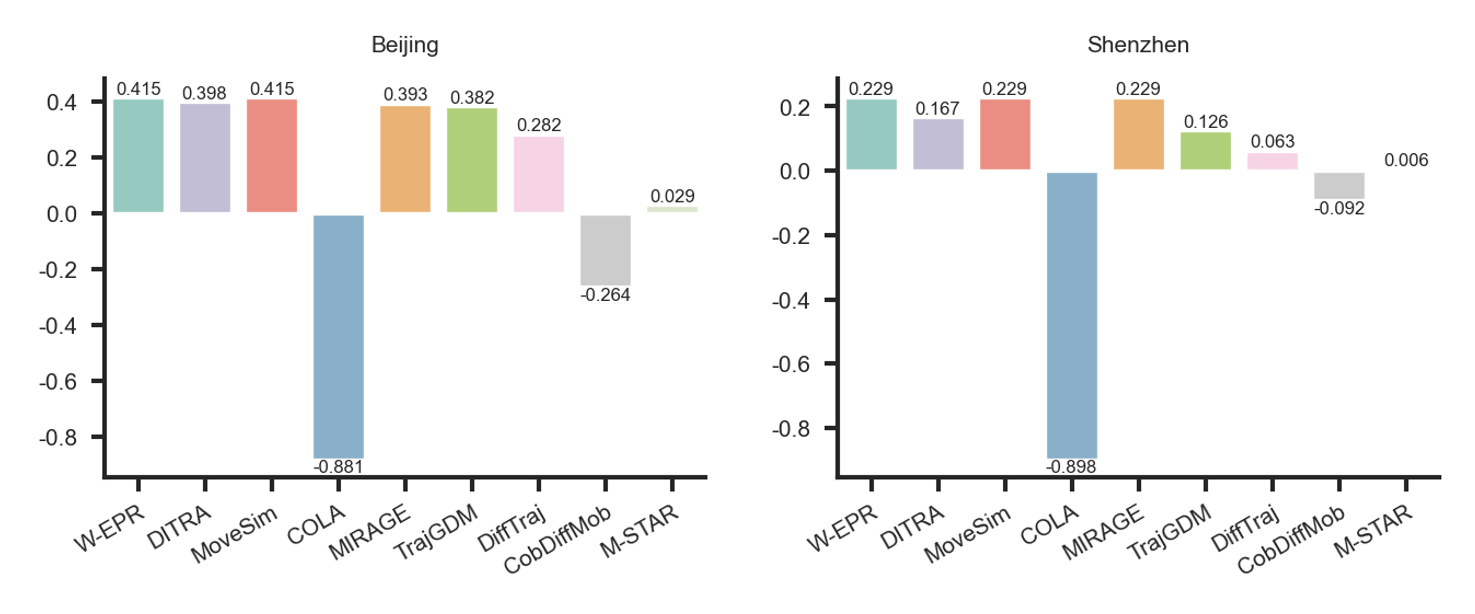}
    \caption{Comparison of trajectory diversity between real and generated datasets. The \textit{Diversity Error} measures the relative difference in the number of unique trajectories between generated and real data.}
    \label{fig:diversity}
\end{figure}
\subsection{Main Results}
\cref{tab:main_results} summarizes the performance of M-STAR and baselines on long-term trajectory generation in Beijing and Shenzhen, evaluated by individual- and population-level metrics. M-STAR consistently outperforms all baselines, ranking first in nearly all metrics and second in one, demonstrating superior spatiotemporal fidelity.

On the individual level, M-STAR achieves the lowest JSD across almost all four metrics in both cities. Compared to the best-performing diffusion model CoDiffMob, M-STAR reduces JSD by an average of $35.2\%$ on spatial metrics including \textit{Distance} and \textit{Radius}. For temporal metrics, including \textit{Duration} and \textit{LocNum}, M-STAR achieves an average improvement of $83.1\%$, demonstrating its superior ability to capture both fine-grained temporal dynamics and consistent long-term mobility behaviors across the entire week. 

On the population level, M-STAR achieves the lowest MAPE on both \textit{Flow} and \textit{Density}. In Shenzhen, which features a more complex urban structure, M-STAR reaches a MAPE of 0.7312 for Flow and 0.3653 for Density, outperforming the best diffusion-based baseline CoDiffMob by $6.9\%$ and $37.5\%$, respectively. To better illustrate M-STAR’s effectiveness in capturing collective mobility dynamics, we visualize the community-level origin-destination flows in Shenzhen (see \cref{fig:od_flow}) and evaluate their similarity using the Common Part of Commuters (CPC) metric \citep{simini2021deep}. M-STAR produces the most realistic flow pattern, closely resembling the real data, and achieves a CPC score of 0.601, significantly higher than all baselines. 

Beyond fidelity, we also assess the diversity of generated trajectories. As shown in \cref{fig:diversity}, most baselines produce trajectory datasets with artificially inflated diversity, leading to positive Diversity Errors, while CoDiffMob is the only baseline with a negative value, indicating underestimation of behavioral variety. In contrast, our model M-STAR achieves values closest to zero, 0.029 for the Beijing dataset and 0.006 for the Shenzhen dataset, indicating the best alignment with real-world diversity. This alignment is essential for ensuring that the synthetic data faithfully reflects the structural complexity of real-world human mobility.

In addition to fidelity, we evaluate generation efficiency using a batch size of 512, measuring inference time (in minutes). M-STAR generates trajectories 15×–30× faster than diffusion-based models like DiffTraj and CoDiffMob, demonstrating the advantage of our autoregressive coarse-to-fine framework in achieving both high fidelity and efficiency, especially for long-range weekly trajectories.

\subsection{Ablation Study}
We conduct ablation studies to evaluate the impact of M-STAR’s core components: multi-spatial scales, multi-temporal scales, movement attributes, and two practical techniques (Token-wise Sampling and Moving-Average Smoothing). We also compare with a baseline autoregressive (AR) model that lacks the coarse-to-fine structure.
As shown in \cref{tab:ablation}, removing any of these components leads to clear performance degradation. The AR model performs worst on spatial metrics such as \textit{Displacement} and \textit{Radius}, highlighting the importance of next-scale prediction. Excluding either spatial or temporal scales causes significant drops in performance, for instance, in Beijing, the JSD of \textit{Radius} increases from 0.0129 to 0.0455 without spatial scales, and the JSD of \textit{Duration} rises from 0.0087 to 0.0373 without temporal scales. Removing movement attributes also consistently reduces overall quality. 
In addition, disabling Token-wise Sampling or Moving-Average Smoothing noticeably impairs temporal alignment. Overall, M-STAR surpasses all its ablated variants and the AR baseline, confirming the effectiveness of its multi-spatiotemporal-scale framework and user-aware design.
\begin{table*}[!tb]
\centering
\caption{Performance comparison of M-STAR against ablation variants on the Beijing and Shenzhen datasets. \textbf{Bold} denotes best results.}
\label{tab:ablation}
\resizebox{0.99\textwidth}{!}{%
\begin{tabular}{l|cccccc|cccccc}
\toprule
Dataset                    & \multicolumn{6}{c|}{Beijing}                                                                                                   & \multicolumn{6}{c}{Shenzhen}                                                                                                   \\ \midrule
\multirow{2}{*}{Metrics}   & \multicolumn{4}{c|}{Individual-level}                                                                   & \multicolumn{2}{c|}{Population-level}         & \multicolumn{4}{c|}{Individual-level}                                                                   & \multicolumn{2}{c}{Population-level}          \\ \cmidrule(l){2-13} 
                           & Displacement    & Radius          & Duration        & \multicolumn{1}{c|}{DailyLoc}          & Flow            & Density      & Displacement    & Radius          & Duration        & \multicolumn{1}{c|}{DailyLoc}          & Flow            & Density      \\ \midrule
AR                         & 0.0040          & 0.2260          & 0.0269          & \multicolumn{1}{c|}{0.0221}          & 0.7597          & 0.7393          & 0.0111          & 0.3298          & 0.0277          & \multicolumn{1}{c|}{0.0258}          & 0.8209          & 0.8088          \\
~w/o Multi-Spatial Scales   & 0.0064          & 0.0455          & 0.0441          & \multicolumn{1}{c|}{0.0953}          & 0.8173          & 0.3198          & 0.0030          & 0.0579          & 0.0316          & \multicolumn{1}{c|}{0.0695}          & 0.7736          & 0.4538          \\
~w/o Multi-Temporal Scales & 0.0005          & 0.0846          & 0.0373          & \multicolumn{1}{c|}{0.0181}          & 0.7847          & 0.2411          & 0.0034          & 0.0318          & 0.0879          & \multicolumn{1}{c|}{0.0165}          & 0.7547          & 0.4453          \\
~w/o Movement Attributes        & 0.0008          & 0.0426          & 0.0248          & \multicolumn{1}{c|}{\textbf{0.0015}} & 0.7265          & 0.2884          & 0.0024          & 0.0528          & 0.0360          & \multicolumn{1}{c|}{0.0052}          & 0.7334          & 0.3674          \\
~w/o Token-wise Sampling &0.0025 &0.0366&0.0199&\multicolumn{1}{c|}{0.0813}&0.7374&0.2447&0.0030 &0.0682 &0.0214 & \multicolumn{1}{c|}{0.0726} &0.7442 &0.3719 \\
~w/o Moving Average &0.0047 &0.0346 &0.0620 & \multicolumn{1}{c|}{0.0375}&0.8239 &0.2444 &0.0025 &0.0251&0.0585&\multicolumn{1}{c|}{0.0026}&0.7353&0.3766 \\
\rowcolor[HTML]{EFEFEF} 
M-STAR                       & \textbf{0.0005} & \textbf{0.0129} & \textbf{0.0087} & \multicolumn{1}{c|}{0.0039}          & \textbf{0.7192} & \textbf{0.2403} & \textbf{0.0012} & \textbf{0.0197} & \textbf{0.0089} & \multicolumn{1}{c|}{\textbf{0.0047}} & \textbf{0.7312} & \textbf{0.3653} \\ 
\bottomrule
\end{tabular}
}
\end{table*}

\begin{table}[t]
\centering
\caption{Comparison of population-level metrics across multiple spatial scales ($s= 2, 4, 8$) on the Shenzhen dataset.}
\label{tab:multi_scale}
\resizebox{0.99\columnwidth}{!}{%
\begin{tabular}{l|cccccc}
\toprule
\multirow{2}{*}{Spatial Scale}            & \multicolumn{2}{c}{s=2}                                   & \multicolumn{2}{c}{s=4}                                   & \multicolumn{2}{c}{s=8}                    \\ \cmidrule(l){2-7} 
                                      & \multicolumn{1}{c}{Flow} & \multicolumn{1}{c}{Density} & \multicolumn{1}{c}{Flow} & \multicolumn{1}{c}{Density} & \multicolumn{1}{c}{Flow} & Density     \\ \midrule
W-EPR                                     & 1.3779                   & 2.6659                         & 3.5515                   & 6.1087                         & 10.8989                  & 11.0408         \\
DITRAS                                     & 1.6588                   & 4.0483                         & 1.3808                   & 3.9653                         & 1.4452                   & 4.6714          \\
Movesim                                    & 1.5298                   & 2.3611                         & 4.1323                   & 3.1793                         & 3.6363                   & 3.8778          \\
COLA                                    & 0.8496                   & \underline{0.2751}                         & 0.8979                   & \underline{0.2864}                         & 0.9041                   & \underline{0.3209}          \\
MIRAGE                                    & 2.1001                   & 6.5413                         & 1.6476                   & 7.5960                         & 1.6549                   & 8.5810          \\
TrajGDM                                  & 1.0708                   & 1.4836                         & 0.9785                   & 1.0295                         & 2.2772                   & 0.7335          \\
DiffTraj                                 & 1.3235                   & 1.9363                         & 1.3948                   & 1.6985                         & 1.2453                   & 1.4881          \\
CoDiffMob                           & \underline{ 0.8481}              &  0.6248                   & \underline{ 0.8298}             &  0.4528                   & \underline{ 0.7454}             & 0.6012    \\
\rowcolor[HTML]{EFEFEF} 
M-STAR                              & \textbf{0.7796}          & \textbf{0.302}                 & \textbf{0.7681}          & \textbf{0.2553}                & \textbf{0.7011}          & \textbf{0.2848} \\ \bottomrule
\end{tabular}
}
\end{table}
\subsection{Multi-Spatial Scales Analysis}
In urban governance, public service planning, and policy analysis, compatibility across different spatial scales is essential. Fine-grained grids facilitate tasks such as traffic regulation and emergency response, while coarse-grained regions support long-term urban planning, resource allocation, and demographic analysis. Therefore, a trajectory generation model must maintain accuracy across multiple spatial resolutions to meet diverse application needs.
As shown in \cref{tab:multi_scale}, we evaluate M-STAR and baseline models across three spatial resolutions ($s=2, 4, 8$), where $s$ denotes the grid size in kilometers (i.e., 2 km, 4 km, and 8 km). M-STAR consistently outperforms all baselines on both \textit{Flow} and \textit{Density} metrics at all spatial scales. 

\begin{table*}[!tb]
\centering
\caption{Performance comparison on \textcolor{red}{weekdays}, where lower results are better. \textbf{Bold} denotes best results; \underline{underline} denotes second-best.}
\label{tab:weekday_results}
\resizebox{0.99\textwidth}{!}{%
\begin{tabular}{l|cccccc|cccccc}
\toprule
Dataset          & \multicolumn{6}{c|}{Beijing} & \multicolumn{6}{c}{Shenzhen} \\
\midrule
\multirow{2}{*}{Metrics} & \multicolumn{4}{c|}{Individual-level} & \multicolumn{2}{c|}{Population-level} & \multicolumn{4}{c|}{Individual-level} & \multicolumn{2}{c}{Population-level} \\
\cmidrule(lr){2-7} \cmidrule(lr){8-13}
 & Displacement & Radius & Duration & \multicolumn{1}{c|}{DailyLoc} & Flow & Density & Displacement & Radius & Duration & \multicolumn{1}{c|}{DailyLoc} & Flow & Density \\
\midrule
W-EPR            & 0.1472 & 0.3865 & 0.1706 & 0.6238 & 1.0373 & 1.5386 & 0.1497 & 0.3456 & 0.2599 & 0.6370 & 0.8593 & 1.0776 \\
DITRAS           & 0.0045 & 0.1663 & 0.0488 & 0.2925 & 1.6963 & 3.5379 & 0.0057 & 0.1021 & 0.0515 & 0.2188 & 1.9642 & 3.4389 \\
Movesim          & 0.0210 & 0.3826 & 0.0686 & 0.4968 & 1.3818 & 1.3734 & 0.4021 & 0.6823 & 0.4059 & 0.6914 & 1.0807 & 2.0338 \\
COLA          & 0.0153 & 0.1924 & 0.0251 & 0.1892 & 0.7702 & \underline{0.3092} & 0.0159 & 0.2751 & 0.0551 & 0.2942 & \underline{0.7613} & \underline{0.3679} \\
MIRAGE         & 0.0097 & 0.0953 & 0.0432 & 0.1934 & 1.9289 & 4.6421 & 0.0071 & 0.0834 & 0.0695 & 0.1470 & 2.6692 & 5.4591 \\
TrajGDM          & 0.0257 & 0.3066 & 0.0253 & 0.1560 & 1.8763 & 4.3786 & 0.0149 & 0.1565 & 0.0251 & 0.0387 & 1.4061 & 2.0755 \\
DiffTraj         & 0.0154 & 0.0326 & 0.0389 & 0.0929 & 1.0997 & 0.9550 & 0.0438 & 0.0392 & 0.1156 & 0.1131 & 1.1494 & 1.7175 \\
CoDiffMob        & \textbf{0.0004} & \underline{0.0195} & \underline{0.0092} & \underline{0.0482} & \underline{0.7530} & 0.4191 & \underline{0.0091} & \underline{0.0234} & \underline{0.0539} & \underline{0.0580} & 0.7985 & 0.6494 \\
\rowcolor[HTML]{EFEFEF} 
M-STAR           & \underline{0.0008} & \textbf{0.0101} & \textbf{0.0079} & \textbf{0.0074} & \textbf{0.7205} & \textbf{0.2362} & \textbf{0.0010} & \textbf{0.0134} & \textbf{0.0059} & \textbf{0.0051} & \textbf{0.7358} & \textbf{0.3171} \\
\bottomrule
\end{tabular}%
}
\end{table*}

\begin{table*}[!tb]
\centering
\caption{Performance comparison on \textcolor{red}{weekends}, where lower results are better. \textbf{Bold} denotes best results; \underline{underline} denotes second-best.}
\label{tab:weekend_results}
\resizebox{0.99\textwidth}{!}{%
\begin{tabular}{l|cccccc|cccccc}
\toprule
Dataset          & \multicolumn{6}{c|}{Beijing} & \multicolumn{6}{c}{Shenzhen} \\
\midrule
\multirow{2}{*}{Metrics} & \multicolumn{4}{c|}{Individual-level} & \multicolumn{2}{c|}{Population-level} & \multicolumn{4}{c|}{Individual-level} & \multicolumn{2}{c}{Population-level} \\
\cmidrule(lr){2-7} \cmidrule(lr){8-13}
 & Displacement & Radius & Duration & \multicolumn{1}{c|}{DailyLoc} & Flow & Density & Displacement & Radius & Duration & \multicolumn{1}{c|}{DailyLoc} & Flow & Density \\
\midrule
W-EPR            & 0.1605 & 0.4582 & 0.1594 & 0.6236 & 1.0015 & 1.6378 & 0.1564 & 0.3911  & 0.2250 & 0.6375 & 0.8772 & 1.2025 \\
DITRAS           & 0.0172 & 0.2217 & 0.0192 & 0.2294 & 1.6768 & 3.2918 & 0.0165 & 0.2562 & 0.0156 & 0.2685 & 2.0454 & 3.5292 \\
Movesim          & 0.0101 & 0.0837 & 0.0544 & 0.0692 & 1.3568 & 1.3543 & 0.4020 & 0.6632 & 0.3753 & 0.6881 & 1.0890 & 2.1094 \\
COLA          & 0.0201 & 0.2765 & 0.0363 & 0.4412 & 0.8187 & 0.3845 & 0.0149 & 0.2247 & 0.0438 & 0.2352 & \underline{0.7585} & \underline{0.3724} \\
MIRAGE          & 0.0082 & \underline{0.0158} & 0.0678 & 0.0657 & 1.9673 & 4.4314 & 0.0065 & \underline{0.0192} & 0.0135 & 0.0315 & 2.8018 & 5.6190 \\
TrajGDM          & 0.0213 & 0.2756 & 0.0202 & 0.1160 & 1.8210 & 4.0028 & 0.0123 & 0.1254 & 0.0138 & 0.0165 & 1.4422 & 2.1321 \\
DiffTraj         & 0.0153 & 0.0476 & 0.0293 & 0.0676 & 1.1102 & 0.9164 & 0.0436 & 0.0605 & 0.0843 & 0.0854 & 1.1865 & 1.7678 \\
CoDiffMob        & \underline{0.0005} & 0.0159 & \textbf{0.0088} & \underline{0.0264} & \underline{0.7258} & \underline{0.3200} & \underline{0.0087} & 0.0194 & \underline{0.0477} & \underline{0.0412} & 0.7698 & 0.5508 \\
\rowcolor[HTML]{EFEFEF} 
M-STAR           & \textbf{0.0005} & \textbf{0.0062} & \underline{0.0150} & \textbf{0.0058} & \textbf{0.7164} & \textbf{0.2505} & \textbf{0.0017} & \textbf{0.0146} & \textbf{0.0139} & \textbf{0.0058} & \textbf{0.7381} & \textbf{0.3495} \\
\bottomrule
\end{tabular}%
}
\end{table*}
\subsection{Performance across Weekdays and Weekends}
To examine M-STAR’s robustness under different temporal scenarios, we evaluate its performance separately on weekday and weekend trajectories using the two datasets. This analysis reflects the variability in human mobility patterns across different day types, where weekdays typically feature more regular commuting behavior, while weekends are characterized by increased randomness and leisure-oriented movements.

We compute all evaluation metrics, including both individual-level and population-level indicators, separately for weekday and weekend subsets of the Shenzhen dataset. As shown in \cref{tab:weekday_results} and \cref{tab:weekend_results}, M-STAR consistently outperforms baseline models across all metrics on both day types, demonstrating its ability to capture diverse spatiotemporal mobility structures. While overall performance slightly declines on weekends, likely due to increased irregular and leisure-driven movements, M-STAR maintains strong accuracy and robustness, indicating better generalization to less predictable mobility patterns.

These results suggest that M-STAR effectively captures both routine and non-routine mobility behaviors, making it suitable for real-world applications involving varying temporal contexts.
\subsection{User Privacy Protection}
\begin{figure}[!tb]
  \centering
  \includegraphics[width=0.99\columnwidth]{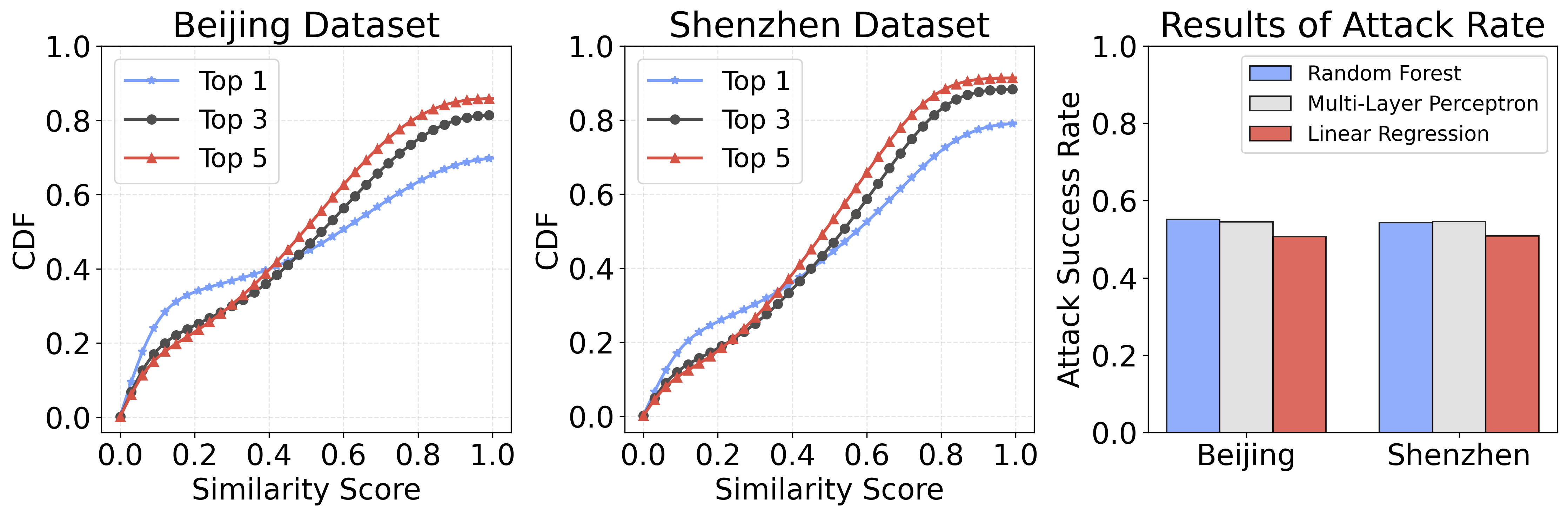}
  \caption{User privacy evaluation: uniqueness testing (left, middle) and MIA success rates (right).}
  \label{fig:privacy}
  \vspace{-0.2cm}
\end{figure}
Due to privacy concerns, real trajectory data usage is often limited. To evaluate potential privacy risks, we conduct two experiments: Uniqueness Testing and Membership Inference Attacks (MIA) \citep{cao2024stage,yuan2024generating}, with results summarized in Figure~\ref{fig:privacy}.

In the uniqueness test, we compare each generated trajectory with real training samples using the Levenshtein distance \citep{lcvenshtcin1966binary}. Most similarity scores fall between 0.3 and 0.8, with none reaching 1, indicating that the generated data captures aggregate patterns without duplicating real samples.

For MIA, following previous work \citep{yuan2024generating}, we adopt an ensemble of three widely used classifiers, Random Forest (RF), Multi-Layer Perceptron (MLP), and Logistic Regression (LR) to conduct membership inference attacks (MIA) \citep{shokri2017membership}. This ensemble approach reduces classifier-specific biases and offers a more comprehensive assessment of membership leakage risk.
Since it is unclear whether a generated trajectory originates from the training (member) set or the test (non-member) set, we adopt a comparative feature construction strategy. For each generated trajectory, we identify its most similar real trajectory (Top-1 match) from both the member and non-member sets using Levenshtein similarity. We then compute a set of trajectory-level features comparing the generated trajectory to each Top-1 match, including Dynamic Time Warping (DTW) distance, cosine similarity, Jaccard similarity, and differences in start and end locations. These features are fed into the classifier, with labels assigned as positive if the matched trajectory is from the member set and negative otherwise. Intuitively, if a generated sample is more similar to a training trajectory, it becomes easier for the classifier to correctly identify it as a member. Attack success rates remain below 0.6, close to the random guess baseline (0.5), indicating limited privacy leakage. 
\subsection{Visualization of Individual Trajectories}
\begin{figure*}[!t] 
  \centering
  \includegraphics[width=0.8\textwidth]{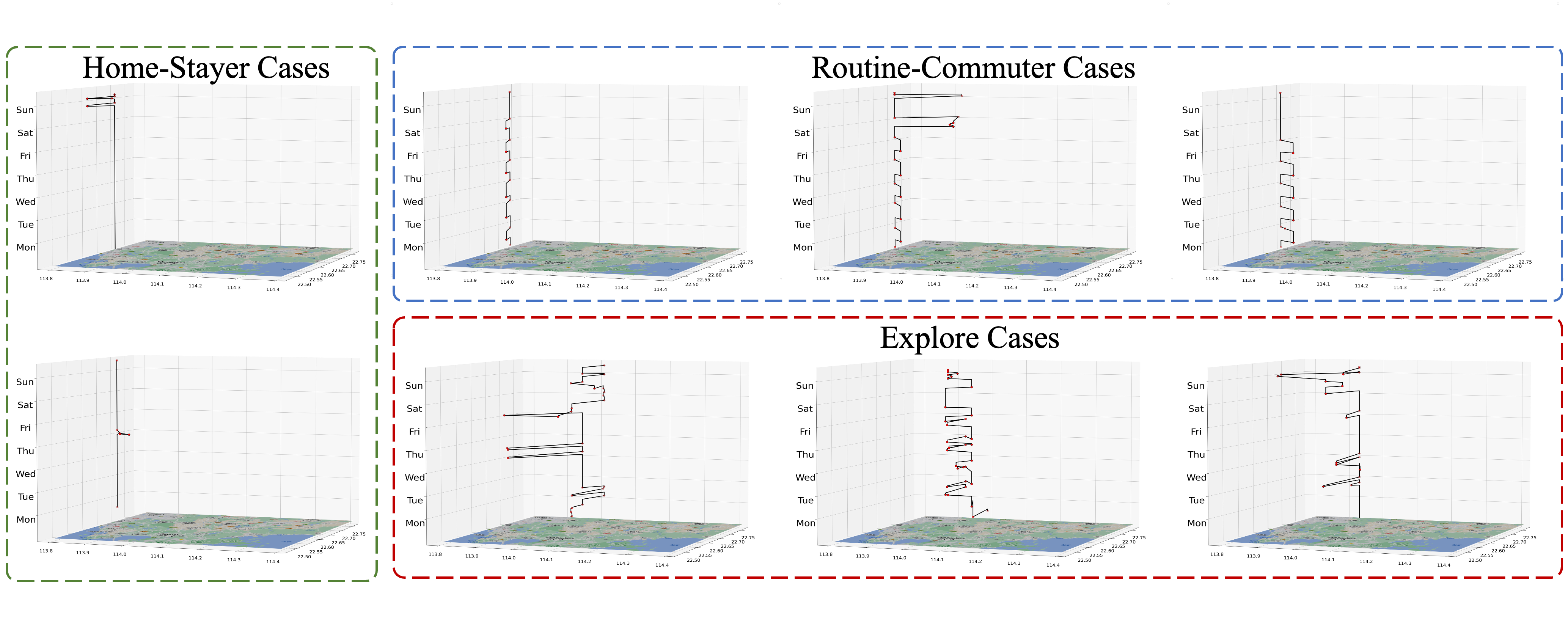}
  \caption{One-week trajectory samples generated by M-STAR, illustrating representative travel patterns: Home Stayer (left), Routine Commuter (top), and Explorer (bottom).}
  \label{fig:individual_traj}
\end{figure*}

To demonstrate the fidelity of generated trajectories, we present examples in \cref{fig:individual_traj} showing one-week trajectories of individuals from the Shenzhen dataset, categorized into three representative movement types: Home Stayer, Routine Commuter, and Explorer. The top three cases illustrate Home Stayers, whose movements are tightly clustered around home locations, a dominant pattern in the dataset. The middle examples represent Routine Commuters, such as students or workers, exhibiting consistent weekday travel between home and commuting locations. The bottom cases show Explorers, characterized by diverse and wide-ranging mobility. Note that with a spatial resolution of 1 km, continuous presence in a grid cell does not necessarily imply staying at home, but rather movement within the same spatial unit.
These visualizations demonstrate M-STAR's ability to generate realistic and behaviorally diverse mobility trajectories that mirror real-world patterns.

\subsection{Hyperparameter Analysis}
\paragraph{\textbf{Spatialtemporal Scales $K$}}
M-STAR adopts a coarse-to-fine framework across multiple spatiotemporal scales. To evaluate the impact of scale granularity, we conduct experiments with different numbers of scales $K \in {2, 4, 8, 12}$ on the Shenzhen dataset. As shown in \cref{tab:hyperparaK}, the number of scales has a significant impact on the spatiotemporal fidelity of generated data. Using only two scales ($K=2$) leads to poor performance across all metrics, indicating insufficient representation capacity. Increasing to $K=4$ greatly improves accuracy, and $K=8$ achieves the best overall performance, striking a balance between temporal coherence and spatial realism. However, further increasing to $K=12$ degrades performance, suggesting that excessive granularity may introduce noise and hamper generalization. These results suggest that moderate scale depth is crucial for effective next-scale prediction.
\begin{table}[!tb]
\centering
\caption{Comparison of spatial-temporal fidelity under various settings of hyperparameter ($K=2, 4, 8, 12$) on the Shenzhen dataset.}
\label{tab:hyperparaK}
\resizebox{0.95\columnwidth}{!}{%
\begin{tabular}{@{}l|cccc|cc@{}}
\toprule
\multirow{2}{*}{Metrics} & \multicolumn{4}{c|}{JSD}                  & \multicolumn{2}{c}{MAPE} \\ \cmidrule(l){2-7} 
                         & Displacement & Radius & Duration & LocNum & Flow      & Population   \\ \midrule
$K=2$                      & 0.0073       & 0.0236 & 0.0617   & 0.0397 & 1.0643    & 1.1487       \\
$K=4$                      & 0.0006       & 0.0209 & 0.0259   & 0.0024 & 0.7235    & 0.2419       \\
$K=8$                    & 0.0005 & 0.0129 & 0.0087   & 0.0039 & 0.7192    & 0.2403       \\
$K=12$                   & 0.0096       & 0.0071 & 0.0685   & 0.0337 & 0.9533    & 0.9020       \\ \bottomrule
\end{tabular}
}
\end{table}
\begin{figure*}[!t] 
  \centering
  \includegraphics[width=0.8\textwidth]{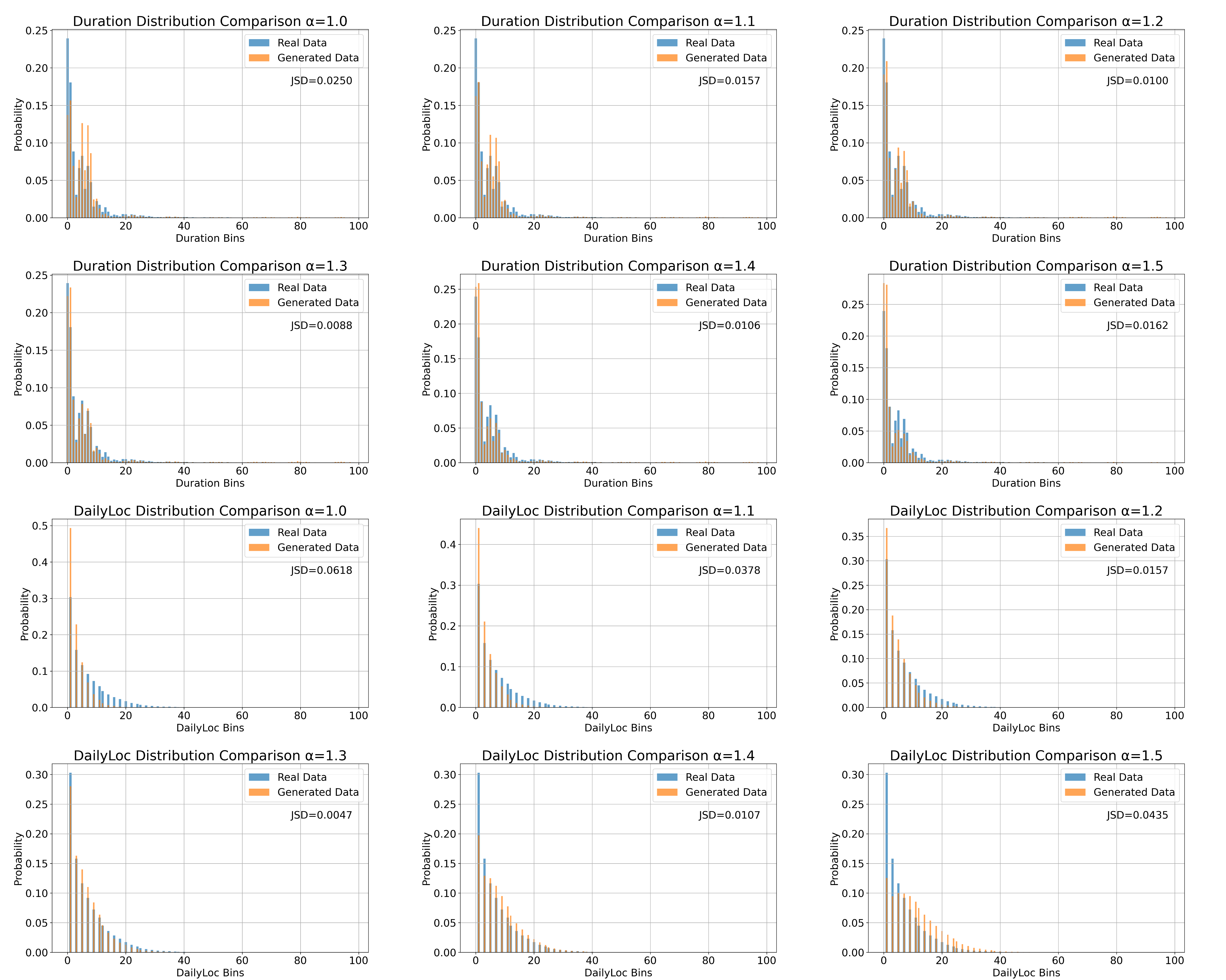}
  \caption{Comparison of the temporal metrics (\textit{Duration} and \textit{DailyLoc}) for M-STAR under various settings of hyperparameter $\alpha$ from 1.0 to 1.5 in on Shenzhen dataset.}
  \label{fig:alpha}
\end{figure*}
\paragraph{\textbf{Token-Wise Sampling $\alpha$}}
The hyperparameter $\alpha$ in Token-Wise Sampling controls the level of randomness during sampling and helps reduce intra-scale temporal discontinuities. To assess its impact, we evaluate temporal alignment metrics (\textit{Duration} and \textit{DailyLoc}) on the Shenzhen dataset with $\alpha$ values ranging from 1.0 to 1.5. As shown in \cref{fig:alpha}, a larger $\alpha$ results in more frequent location changes and shorter visit durations, introducing higher variability in the generated trajectories. In contrast, a smaller $\alpha$ leads to lower randomness and overly deterministic patterns. A moderate value (e.g., $\alpha=1.3$) yields the best temporal alignment. This demonstrates that careful tuning of $\alpha$ is essential to balance diversity and temporal consistency in generated trajectories. 

\subsection{Utility of the Generated Trajectories}
\paragraph{\textbf{Next-Location Prediction}} To assess the practical utility of generated trajectories, we conduct a next-location prediction task using the Shenzhen dataset. An LSTM-based model predicting the next three locations is trained on three datasets: real trajectories, M-STAR generated, and CobDiffMob generated, and evaluated on the test set from the real dataset. As shown in \cref{tab:application_prediction}, both M-STAR and CobDiffMob achieve strong performance across the Accuracy metrics, with M-STAR performing slightly closer to the real data. 
\begin{table}[h]
\centering
\caption{Accuracy of next-location prediction based on generated and real data.}
\label{tab:application_prediction}
\small
\begin{tabular}{l|cccc}
\toprule
Method & ACC     & $\text{ACC}_{\text{t+1}}$  & $\text{ACC}_{\text{t+2}}$    & $\text{ACC}_{\text{t+3}}$   \\
\midrule
Real & 0.8765  & 0.9305 & 0.8727 & 0.8264   \\
CobDiffMob & 0.8671 & 0.9247 &  0.8628 & 0.8137 \\
\rowcolor[HTML]{EFEFEF} 
M-STAR & 0.8742 & 0.9291 & 0.8701 & 0.8235 \\ 
\bottomrule
 \end{tabular}
\end{table}
\paragraph{\textbf{Epidemic Simulation}}
Synthetic trajectories are valuable for epidemic modeling. We evaluate their effectiveness using a Susceptible–Exposed–Infectious–Recovered (SEIR) compartmental model. Assuming a basic reproduction number $R_0 = 2.0$ (close to the original strain of SARS-CoV-2 ), we construct hourly contact networks and simulate disease spread over 10 weeks using 10,000 sampled trajectories from the Shenzhen dataset, repeated across 10 runs. As shown in \cref{tab:application_ep}, M-STAR consistently achieves lower MAPE than CobDiffMob when comparing simulated epidemic outcomes with real-data-based simulations across all three compartments (Exposed, Infectious, Recovered), indicating stronger preservation of critical mobility structure.
\begin{table}[h]
\centering
\caption{Mean Absolute Percentage Error (MAPE) between generated and real data in epidemic simulations.}
\small
\setlength{\tabcolsep}{3pt}
\label{tab:application_ep}
\begin{tabular}{l|cccc}
\toprule
Method    &E (Active) & I (Active) & R (Active) & I (Cumulative)   \\ \midrule
CobDiffMob & 0.2167         & 0.1573            & 0.1483    &  0.1294      \\
\rowcolor[HTML]{EFEFEF} 
M-STAR     & 0.1078         & 0.0626            & 0.0994     & 0.0689      \\ \bottomrule
\end{tabular}
\end{table}

\section{Related Work}
\subsection{Synthetic Trajectory Generation}
Trajectory generation methods fall into two main categories: mechanism-driven and deep learning-based approaches. Mechanism-driven models, such as the Exploration and Preferential Return (EPR) model \citep{song2010modelling}, simulate human mobility through interpretable rules like exploration and return behavior but are constrained by strong assumptions that limit their expressiveness \citep{barbosa2015effect,jiang2016timegeo}. In contrast, deep learning methods, including Generative Adversarial Networks (GANs) \citep{feng2020learning,yuan2022activity}, Variational Autoencoders (VAEs) \citep{huang2019variational,long2023practical}, and more recently diffusion models offer greater flexibility for modeling complex spatiotemporal dynamics. For instance, MoveSim \citep{feng2020learning} leverages SeqGAN \citep{yu2017seqgan} to learn sequential mobility patterns, while VOLUNTEER \citep{long2023practical} employs a two-layer VAE to model user preferences and actions. Building on the success of diffusion models in image synthesis \citep{song2020denoising,peebles2023scalable}, recent works like DiffTraj \citep{zhu2023difftraj} and TrajGDM \citep{chu2024simulating} first applied them to trajectory generation, enhancing quality through iterative denoising. Diff-RNTraj \citep{wei2024diff} and CobDiffMob \citep{zhang2025noise} further incorporate road networks as spatial constraints, a design well-suited for vehicle trajectory generation where road geometry and driving dynamics play a central role. Moreover, CobDiffMob \citep{zhang2025noise} integrated collaborative noise priors with the OD-based EPR model to model both individual movement characteristics and population-wide dynamics. However, these methods often suffer from high computational costs, particularly for long-term trajectory generation, and generally overlook the explicit modeling of multi-scale spatiotemporal structures that are essential for capturing hierarchical mobility dynamics.

\subsection{Coarse-to-fine Autoregressive Models}
Building on the success of next-token autoregressive models in language modeling \citep{touvron2023llama,achiam2023gpt} and visual autoregressive frameworks like VQVAE in image synthesis \citep{van2017neural}, \citet{tian2024visual} introduced a coarse-to-fine next-scale prediction approach that significantly improves image generation quality and scalability. This autoregressive coarse-to-fine paradigm was subsequently adopted in various domains. For example, CARP reformulated autoregressive action generation as a coarse-to-fine process for visuomotor policy learning \citep{gong2022diffuseq}; SAT compressed audio sequences into discrete tokens and synthesized them autoregressively for efficient audio generation \citep{qiu2024efficient}; and VARSR applied the next-scale framework to image super-resolution, achieving state-of-the-art performance on multiple quality metrics while improving generation efficiency by a factor of ten compared to diffusion-based methods \citep{qu2025visual}. However, applying this coarse-to-fine paradigm to human trajectory generation is challenging. Unlike images or temporal signals \citep{tian2024visual,van2017neural}, trajectories are sequences of discrete spatiotemporal units, making standard multiscale operations such as downsampling or pooling inapplicable. Moreover, spatial and temporal scales are tightly coupled, so scale transitions must capture both dimensions rather than spatial refinements alone. These differences limit the direct use of existing next-scale pipelines and motivate a specialized coarse-to-fine autoregressive design for mobility data.
\section{Conclusion}

This study presents M-STAR, a multi-scale spatiotemporal autoregressive framework for human trajectory generation. It combines a Multi-scale Spatiotemporal Tokenizer and STAR-Transformer to generate trajectories in a coarse-to-fine manner, guided by individual movement attributes. Experiments on real-world datasets demonstrate that M-STAR outperforms diffusion-based and mechanistic baselines across both individual- and population-level metrics, while also delivering significantly faster generation. Privacy evaluations show low risk of data leakage, and the generated trajectories support downstream tasks such as next-location prediction and epidemic simulation. Future work will focus on enhancing the cross-city generalization of M-STAR by reducing its reliance on city-specific spatiotemporal patterns through adaptive scaling or retraining strategies. Overall, M-STAR provides a practical and scalable solution for various applications such as urban planning, epidemic modeling, and public policy analysis.


\ifCLASSOPTIONcaptionsoff
  \newpage
\fi

\bibliographystyle{IEEEtranN}
\bibliography{papers}
\begin{IEEEbiography}[{\includegraphics[width=1in,height=1.25in,clip,keepaspectratio]{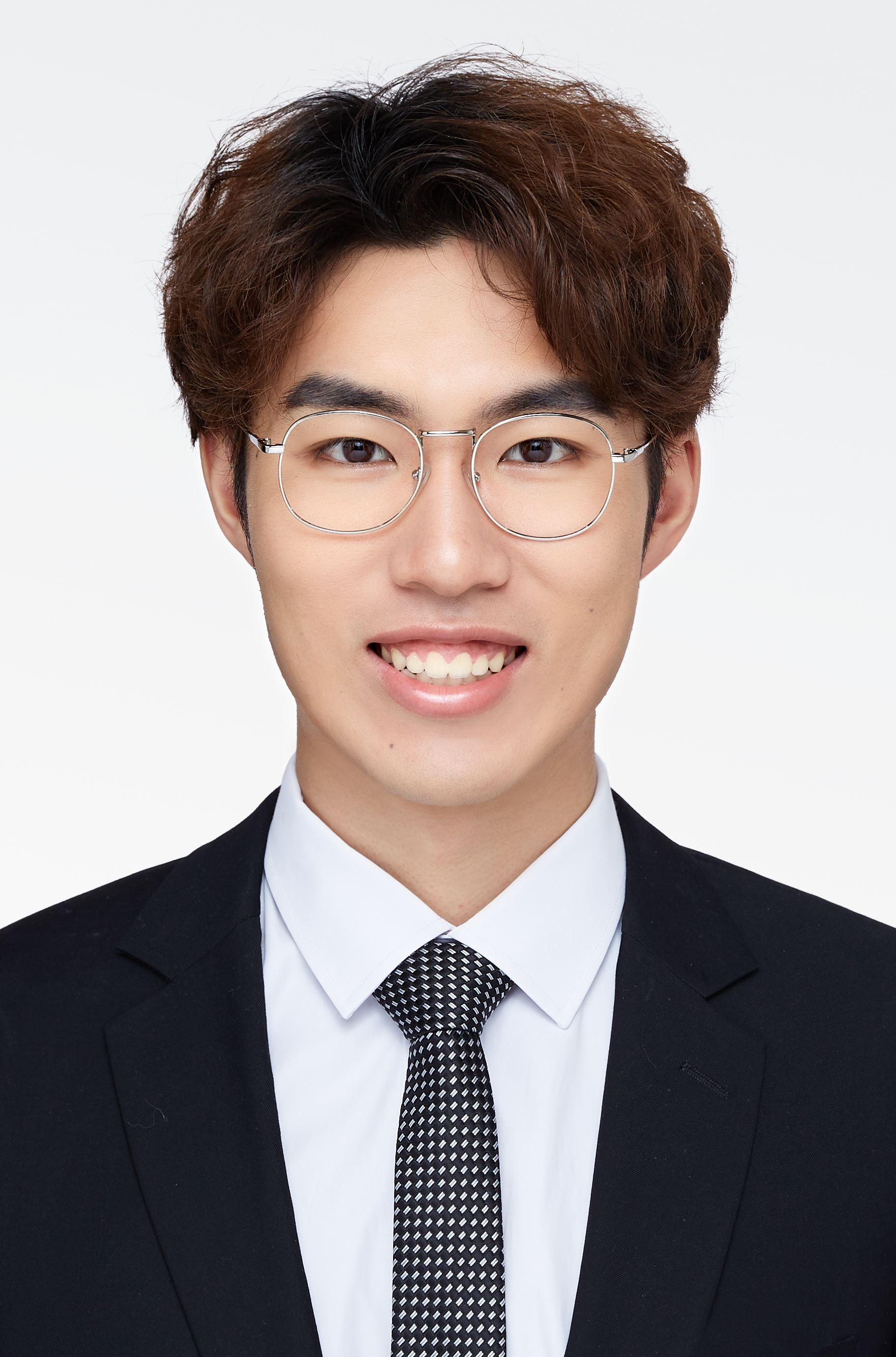}}]{Yuxiao Luo}
received the B.E. degree from the University of Electronic Science and Technology of China (UESTC), Chengdu, China. 
He is currently pursuing the Ph.D. degree with The Hong Kong Polytechnic University, Hong Kong, in collaboration with the 
Shenzhen Institutes of Advanced Technology (SIAT), Chinese Academy of Sciences (CAS). 
His research interests include spatiotemporal data mining and human mobility modeling. 
\end{IEEEbiography}
\begin{IEEEbiography}[{\includegraphics[width=1in,height=1.25in,clip,keepaspectratio]{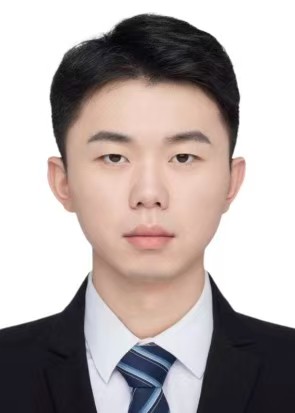}}]{Songming Zhang}
received the M.S. degree from Chongqing Jiaotong University, Chongqing, China. He is currently pursuing the Ph.D. degree at Nanjing University, Nanjing, China. His research interests include trustworthy machine learning and AI for science. 
\end{IEEEbiography}
\begin{IEEEbiography}[{\includegraphics[width=1in,height=1.25in,clip,keepaspectratio]{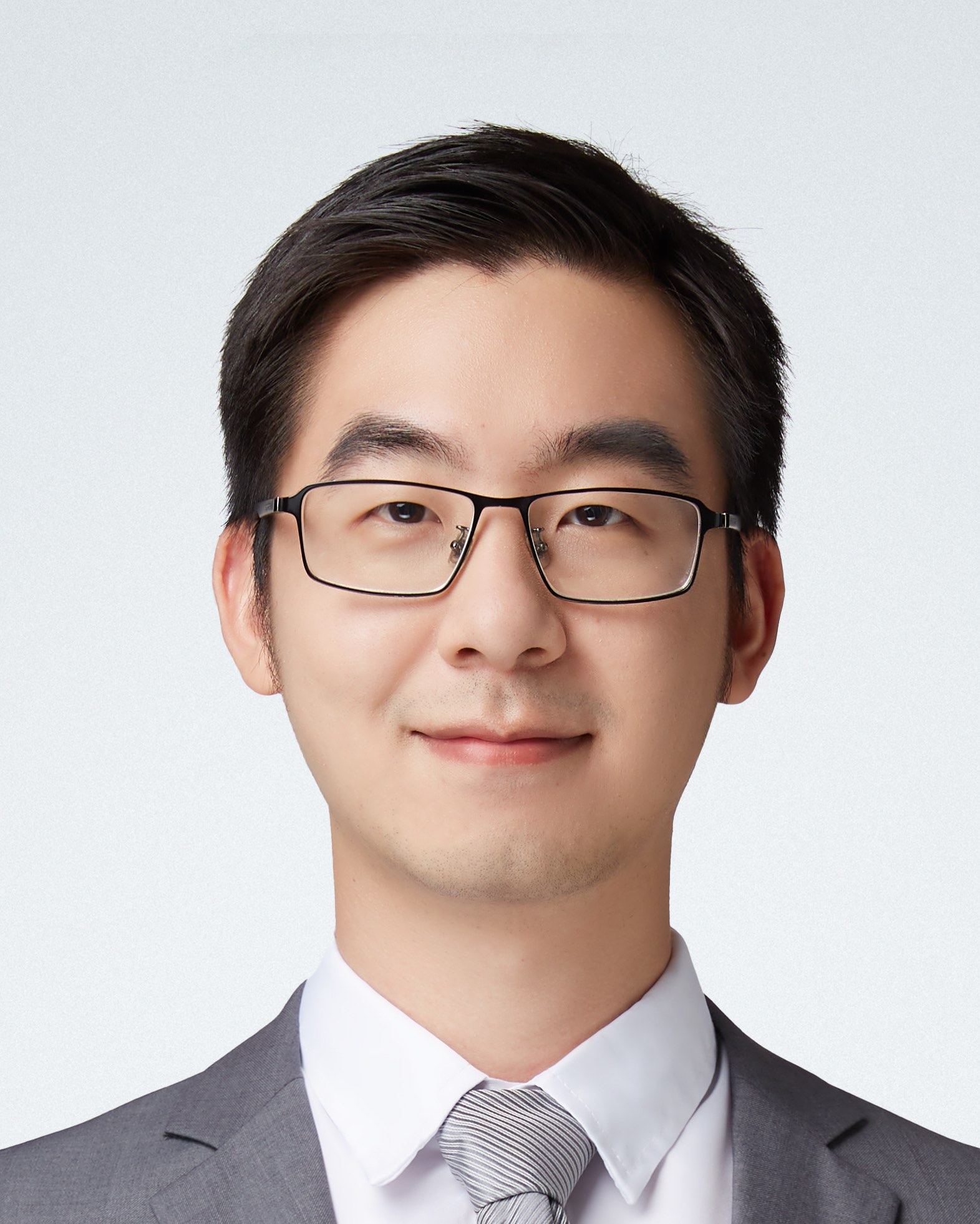}}]{Sijie Ruan}
(Member, IEEE) received the B.E. and Ph.D. degrees from Xidian University, Xian, China, in 2017 and 2022, respectively. 
He is currently an Assistant Professor with the School of Computer Science and Technology, Beijing Institute of Technology. 
His research interests include spatio-temporal data mining and urban computing.
\end{IEEEbiography}
\begin{IEEEbiography}[{\includegraphics[width=1in,height=1.25in,clip,keepaspectratio]{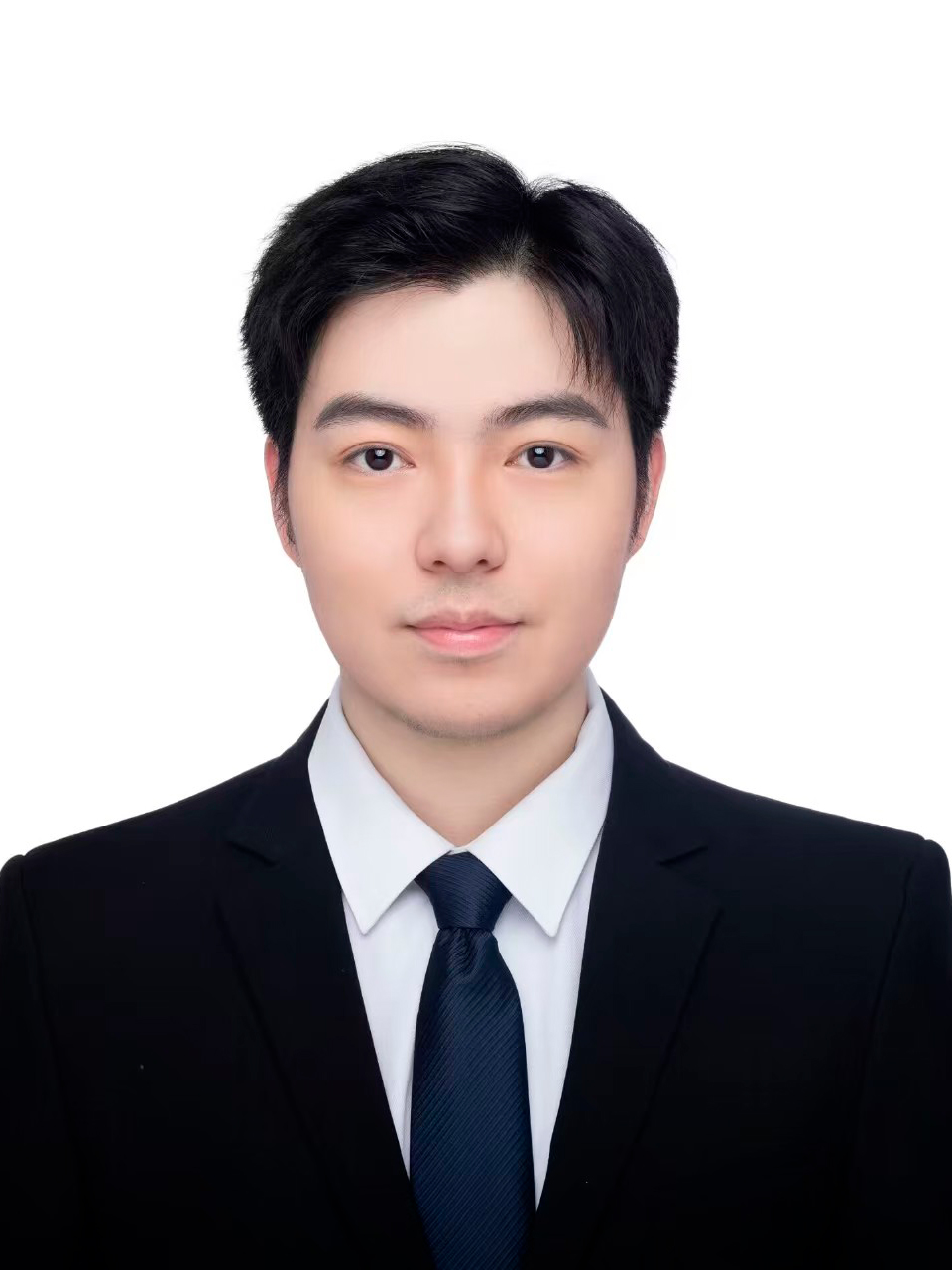}}]{Siran Chen}
received the B.E. degree from the University of Chinese Academy of Sciences, Beijing, China. 
He is currently pursuing the Ph.D. degree with the Shenzhen
Institute of Advanced Technology (SIAT), Chinese
Academy of Sciences (CAS). 
His research interests include video understanding and multimodal large language models.
\end{IEEEbiography}
\begin{IEEEbiography}[{\includegraphics[width=1in,height=1.25in,clip,keepaspectratio]{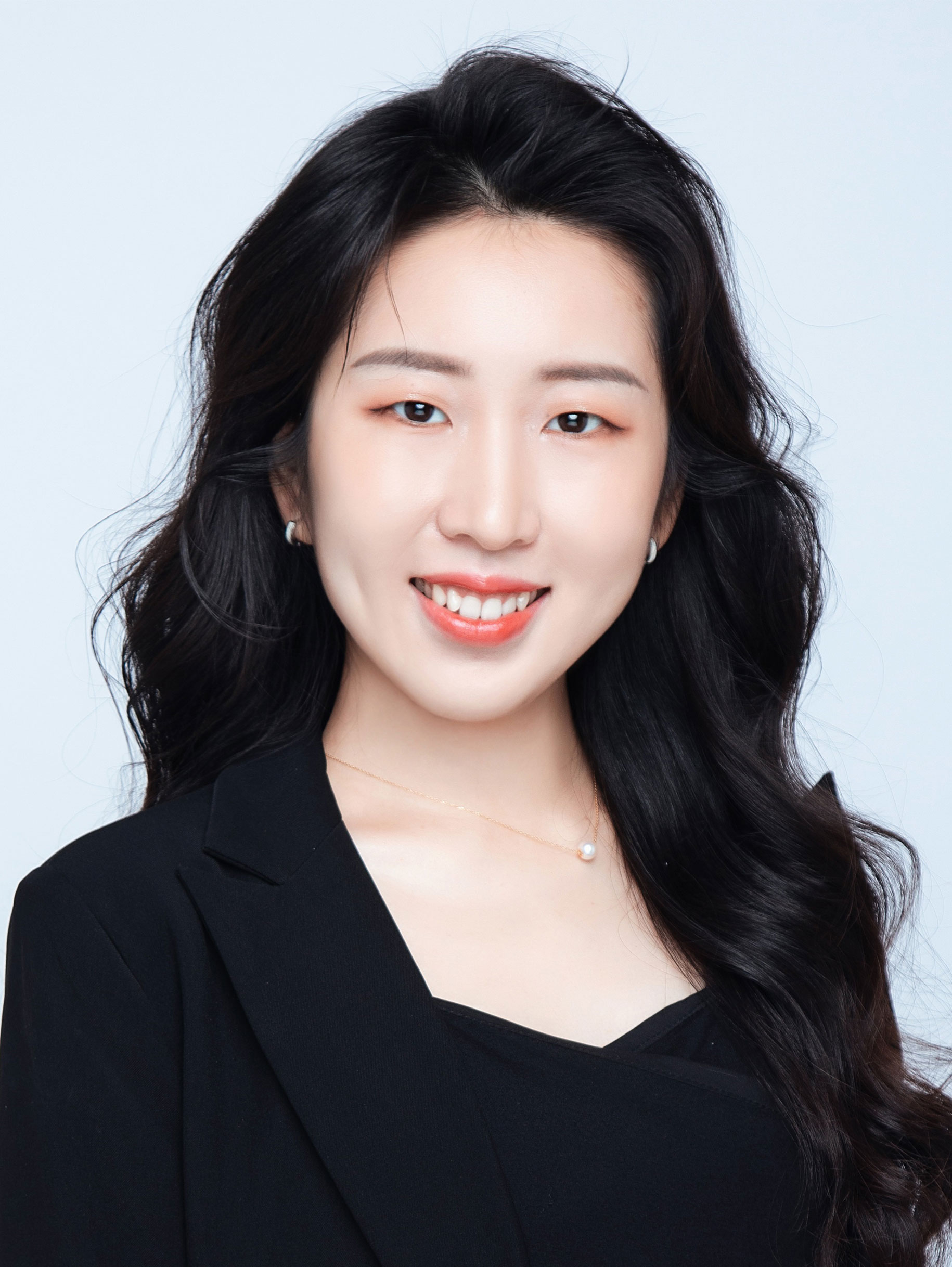}}]{Kang Liu}
received the B.E. degree from China University of Geosciences, Beijing, China, 
and the Ph.D. degree from the Institute of Geographic Sciences and Natural Resources Research, 
Chinese Academy of Sciences (CAS), Beijing, China. She is currently an Associate Professor 
with the Shenzhen Institute of Advanced Technology (SIAT), CAS. Her research interests 
include human mobility modeling and geospatial artificial intelligence (GeoAI).
\end{IEEEbiography}
\begin{IEEEbiography}[{\includegraphics[width=1in,height=1.25in,clip,keepaspectratio]{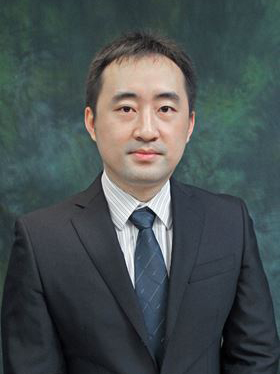}}]{Yang Xu}
received the B.S. degree in remote sensing and photogrammetry and the M.S. degree in geographic information science from Wuhan University in 2009 and 2011, respectively, and the Ph.D. degree in geography from the University of Tennessee.Knoxville, in 2015. He is currently an Associate 
Professor with the Department of Land Surveying and Geo-Informatics, The Hong Kong 
Polytechnic University, Hong Kong. His research interests include human mobility, 
GIScience and urban informatics.
\end{IEEEbiography}
\begin{IEEEbiography}[{\includegraphics[width=1in,height=1.25in,clip,keepaspectratio]{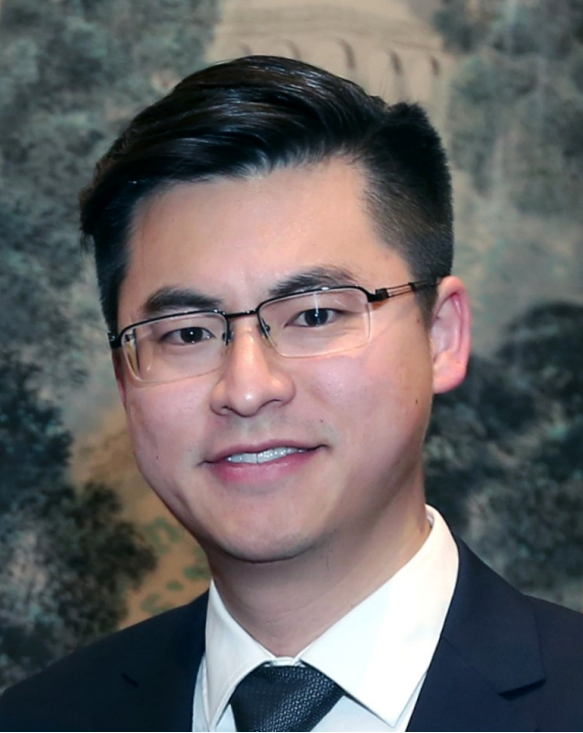}}]{Yu Zheng}
(Fellow, IEEE) is the Vice President of JD.com and the President of JD Intelligent Cities Research. 
Before joining JD.com, he was a Senior Research Manager with Microsoft Research. 
He is also a Chair Professor with Shanghai Jiao Tong University. He served as the 
Editor-in-Chief of ACM Transactions on Intelligent Systems and Technology and 
as Program Co-Chair of ICDE 2014 and CIKM 2017. He delivered keynote speeches 
at AAAI 2019, KDD 2019 Plenary Keynote Panel, and IJCAI 2019 Industrial Days. 
He received the SIGKDD Test-of-Time Award twice (in 2023 and 2024). He was 
named one of the Top Innovators Under 35 by MIT Technology Review (TR35) and 
is an ACM Distinguished Scientist, recognized for his contributions to spatio-temporal 
data mining and urban computing.
\end{IEEEbiography}
\begin{IEEEbiography}[{\includegraphics[width=1in,height=1.25in,clip,keepaspectratio]{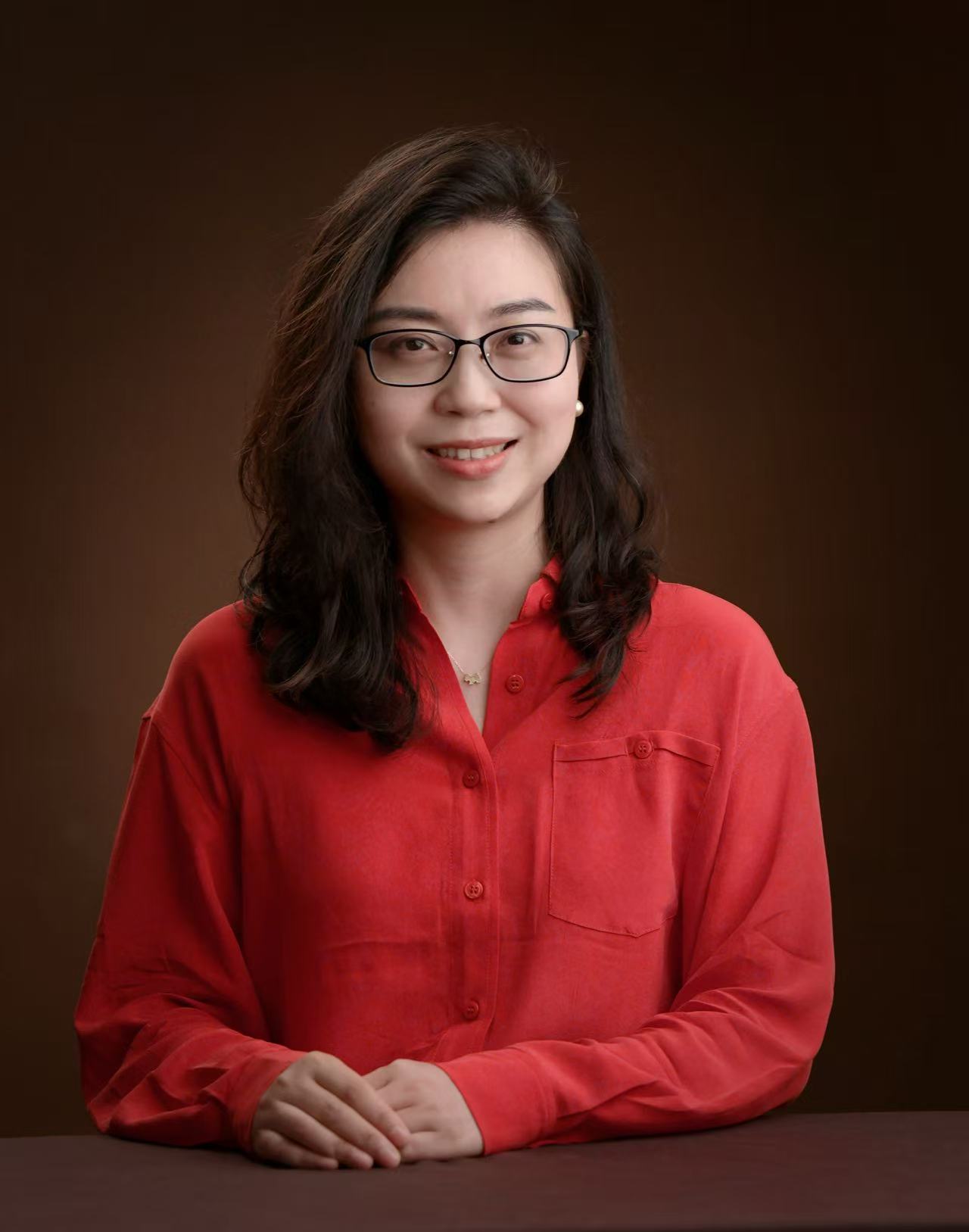}}]{Ling Yin}
received the B.E. and M.S. degree from Nanjing
University, Nanjing, China, and the Ph.D. degree
from the University of Tennessee, Knoxville, TN,
USA. She is currently a Professor with the Shenzhen
Institute of Advanced Technology (SIAT), Chinese
Academy of Sciences (CAS), and an Adjunct Professor
 with the Faculty of Computer Science and
Control Engineering, Shenzhen University of Advanced
 Technology. Her research interests include
spatiotemporal data intelligence, human mobility,
and spatiotemporal epidemic modelling.
\end{IEEEbiography}


\end{document}